\documentclass[runningheads]{llncs}

\usepackage{eccv}

\usepackage{eccvabbrv}

\usepackage{graphicx}
\usepackage{booktabs}

\usepackage[accsupp]{axessibility}  

\usepackage{hyperref}       
\usepackage[utf8]{inputenc} 
\usepackage[T1]{fontenc}    
\usepackage{url}            
\usepackage{booktabs}       
\usepackage{amsfonts}       
\usepackage{nicefrac}       
\usepackage{microtype}      
\usepackage{xcolor}         

\usepackage{graphicx}
\usepackage{multirow, multicol}
\usepackage{mathtools}
\usepackage{bbm}
\usepackage{mathabx}
\usepackage{algorithm}
\usepackage{algorithmic}
\usepackage{colortbl}
\usepackage{amssymb}
\definecolor{mygray}{gray}{0.9}
\usepackage{wrapfig} 
\usepackage{caption}
\usepackage{pifont}
\usepackage{subcaption}
\usepackage{enumitem}
\usepackage{soul}
\usepackage[table]{xcolor}
\usepackage{orcidlink}

\newcommand{\vllm}{{Video-LLMs}}
\newcommand{\snow}{\includegraphics[height=1em, trim=5 5 5 5, clip]{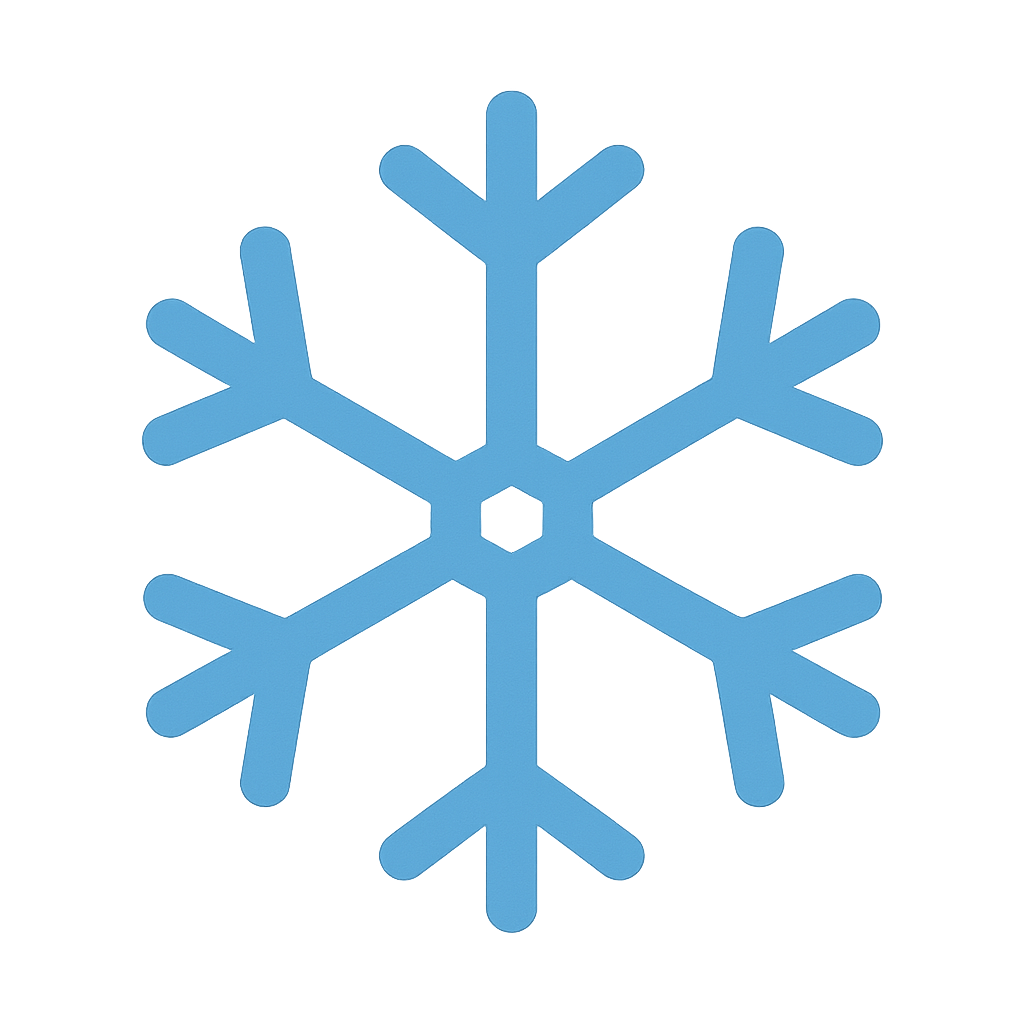}}
\newcommand{\fire}{\includegraphics[height=1em, trim=5 5 5 5, clip]{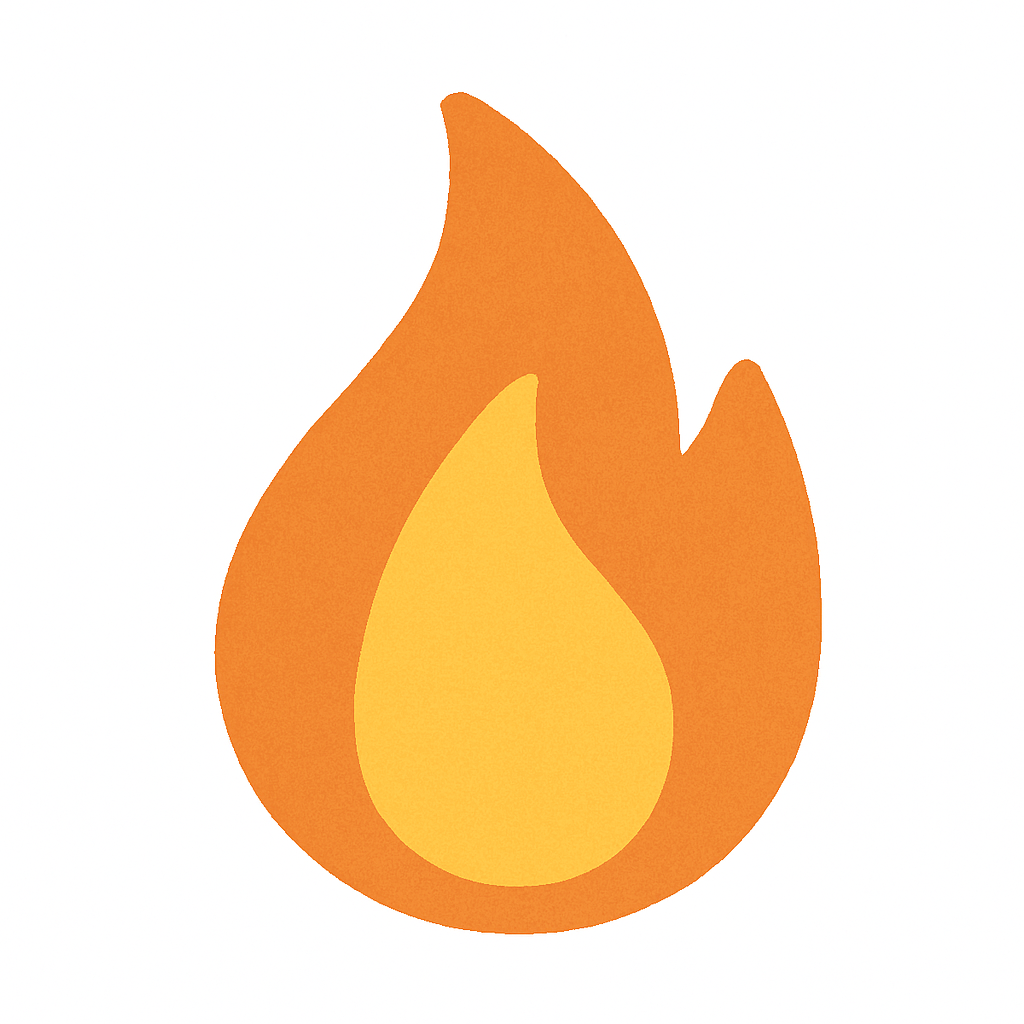}}

\usepackage{soul}

\begin{document}

\title{EgoExo-Con: Exploring View-Invariant \\ Video Temporal Understanding} 

\titlerunning{M.Jung et. al}

\author{
Minjoon Jung\inst{1} \and
Junbin Xiao\inst{2,3}\thanks{Corresponding Author} \and
Junghyun Kim\inst{1} \and \\
Byoung-Tak Zhang\inst{1} \and
Angela Yao\inst{3}
}

\authorrunning{M.Jung et al.}

\institute{
Seoul National University \and
University of Science and Technology of China \and
National University of Singapore \\
\email{\{mjjung, jhkim, btzhang\}@bi.snu.ac.kr},
\email{junbinxiao@ustc.edu.cn}, 
\email{ayao@comp.nus.edu.sg}
}

\maketitle
\begin{abstract}
Do Video-LLMs have \emph{consistent} temporal understanding when videos capture the same event from different viewpoints? To study this question, we introduce EgoExo-Con(sistency), a benchmark of synchronized egocentric and exocentric video pairs with human-refined queries that ensure all concepts are visible in both viewpoints. EgoExo-Con emphasizes two temporal understanding tasks: Temporal Verification and Temporal Grounding. It evaluates not only correctness but consistency across viewpoints. Our analysis reveals two critical limitations of existing Video-LLMs: (1) models often fail to maintain consistency, with results far worse than their single-view performances. (2) When naively finetuned with synchronized videos of both viewpoints, the models show improved consistency but often underperform those trained on a single view. For improvements, we propose View-GRPO, a novel reinforcement learning framework that effectively strengthens view-specific temporal reasoning while encouraging consistent comprehension across viewpoints. Our method demonstrates its superior temporal understanding capabilities, especially for improving cross-view consistency. All resources have been made available at \href{https://github.com/minjoong507/EgoExo-Con}{EgoExo-Con}.
\end{abstract}
\section{Introduction}
\begin{figure}[t]
        \centering
        \includegraphics[width=1\linewidth]{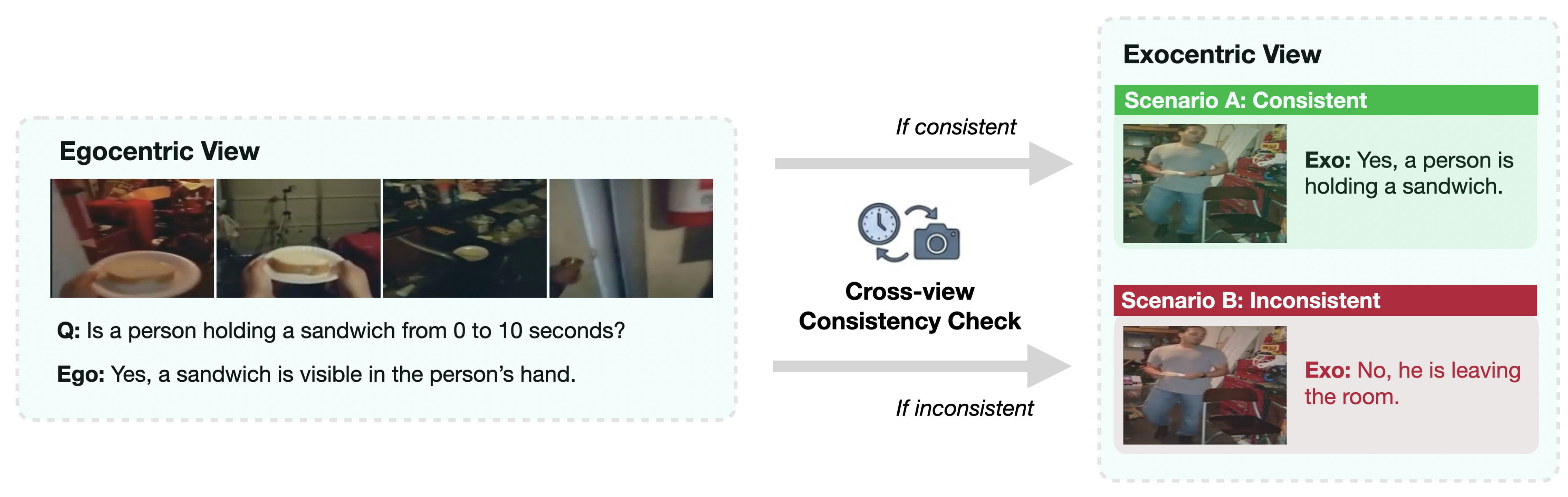}
        \vspace{-4mm}
    \caption{
    \textbf{Evaluation setup in EgoExo-Con.} The same query is applied to synchronized egocentric and exocentric video pairs, and model responses are checked for consistency.
    }
    \label{fig: teaser}
    \vspace{-5mm}
\end{figure}
Recent advances in video large language models (Video-LLMs) have shown impressive capabilities in question answering \cite{zhang2024video,wang2024videoagent,wang2025internvl3} and temporal grounding \cite{timechat,vtimellm,zeng2024timesuite,wang2025time}, and are stepping towards fine-grained and long-range reasoning \cite{zhang2024long,shen2024longvu, wang2025internvideo2,feng2025videor1}. However, most benchmarks \cite{mangalam2023egoschema,fu2024videomme,wu2024longvideobench,zhou2025mlvu} and methods assume a fixed or minimally varying viewpoint, \eg~third-person view (exo) videos. This begs a critical question: \emph{Do Video-LLMs achieve consistent temporal understanding across different camera perspectives?} 

Often, videos of the same event appear strikingly different when captured from different perspectives. A cooking demonstration filmed from a head-mounted camera (ego view) looks unlike a side-mounted tripod shot (exo view). Yet, the underlying temporal dynamics, such as cutting vegetables and stirring a pot, are identical. For humans, this view variation rarely impedes understanding; we easily track the sequence of actions and localize their temporal moments across viewpoints. This makes temporal reasoning particularly critical: while appearance cues can vary drastically with viewpoint, the temporal structure of events is invariant. Thus, evaluation of cross-view consistency is essential and can be effectively carried out through temporal understanding tasks; however, such capabilities remain largely underexplored in current \vllm.

To study this, we introduce EgoExo-Con, a benchmark comprising 1,148 synchronized videos and 2,269 human-refined temporal-bounded event queries, to evaluate whether models can provide consistent predictions across viewpoints - a key indicator of view-invariant video-language understanding. The benchmark focuses on two temporal understanding tasks: \emph{temporal verification}~\cite{jung2025consistency} and \emph{temporal grounding}~\cite{gao2017tall}. Temporal verification is a question-answering (QA) task that asks whether a given event occurs within a specific video moment, while temporal grounding requires identifying the relevant video moment (start and end timestamps) corresponding to an event query. As shown in Fig.~\ref{fig: teaser}, we ask the same event but with synchronized videos of different viewpoints, and check if the tested models can output correct and consistent responses.

We evaluate both the advanced closed-source models \cite{comanici2025gemini, 2025gpt5} and open-source \vllm~comprising general-purpose \cite{li2024mvbench,zhang2025videollama3,bai2025qwen2, cheng2024videollama2} and time-aware variants \cite{timechat,vtimellm,zeng2024timesuite,jung2025consistency}. Our results reveal that all models, especially the open-source ones, struggle with cross-view consistency. They generally exhibit a modest performance gap between individual ego and exo videos, but achieve consistency scores barely over half their single-view performance in both tasks. This indicates that the relatively stable performances across viewpoints may be sourced from view-specific biases rather than robust cross-view temporal understanding. 

Our further investigations show that naive multi-view supervised fine-tuning (SFT) with synchronized video-language data is insufficient.  In fact, it even underperforms the counterpart with single-view training, indicating that naively merging viewpoints may introduce conflicting priors that undermine temporal signals and consistency. Furthermore, the improvements are unstable and ineffective even when the visual encoder is unfrozen during training. These collectively suggest that \emph{viewpoint variation remains a significant challenge for current \vllm~in robust video temporal understanding}. 

To that end, we propose View-GRPO, a reinforcement learning (RL) framework for view-invariant video temporal understanding. It encourages the model to learn consistent cross-view temporal reasoning as well as certain view-specific details. Experiments demonstrate that View-GRPO yields more robust and consistent video understanding on our benchmark and generalizes well to other video understanding benchmarks. In summary, EgoExo-Con establishes a new paradigm for evaluating and improving view-invariant temporal understanding in Video-LLMs. We hope it will foster future research on models that truly capture the essence of dynamic events independent of perspective. Our primary contributions are as follows:
\begin{itemize}[leftmargin=*]
    \item We study the robustness of \vllm~in cross-view video temporal understanding and introduce EgoExo-Con, a synchronized ego–exo benchmark constructed with manual annotation efforts.
   \item We reveal that current \vllm~achieve cross-view consistency barely better than half of their single-view performance, and naively blending perspectives for training could introduce conflicting priors, undermining consistency.
    \item We propose View-GRPO, a reinforced approach that explicitly strengthens temporal reasoning while encouraging view-invariant comprehension, significantly improving robust and consistent video temporal understanding.
\end{itemize}
\section{Related Work}
\paragraph{\bf Video Large Language Models.}
\vllm~\cite{li2024mvbench, zhang2024video, zhang2025videollama3, bai2025qwen2, wang2025internvl3, comanici2025gemini} integrate pretrained video representations into powerful LLMs \cite{grattafiori2024llama, bai2025qwen2,yang2025qwen3} to enable chatting about videos. While advances to date are mostly achieved in short and coarse-grained question answering \cite{xu2017video,yu2019activitynet}, more recent \vllm~\cite{timechat, vtimellm, qian2024momentor, guo2024vtgllm, wang2024hawkeye, zeng2024timesuite,meinardus2024chrono,jung2025consistency, li2025universal} have explored grasping fine-grained temporal moments (``when''). However, all of these models solve videos captured in a single camera viewpoint (either exo or ego) and do not evaluate whether temporal reasoning remains stable across views of the same event. This work thus fills such gap by conducting a comprehensive analysis.

\paragraph{\bf Ego-Exo Benchmarks.}
Most benchmarks target either exocentric \cite{gao2017tall,yu2019activitynet,xiao2021next,xiao2024can,fu2024videomme} or egocentric \cite{mangalam2023egoschema,di2024grounded,cheng2024egothink,ye2025mmego,xiao2025egoblind} video understanding, with only a few offering paired views: CharadesEgo \cite{sigurdsson2018charadesego}, LEMMA \cite{jia2020lemma}, EgoExo-4D \cite{grauman2024egoexo}, Assembly101 \cite{sener2022assembly101}, EgoExo-Fitness \cite{li2024egoexofit}, and EgoExOR \cite{ozsoy2025egoexor}. Yet, they are either domain-specific or do not evaluate cross-view temporal reasoning. A concurrent effort, EgoExoBench \cite{he2025egoexobench}, also explores Video-LMMs in cross-view temporal reasoning, but it primarily targets action ordering via multi-choice selection. While it highlights viewpoint gaps, it neither explicitly addresses prediction consistency nor proposes a concrete methodology. In contrast, we propose EgoExo-Con to evaluate temporal understanding and prediction consistency across synchronized viewpoints with View-GRPO for improvements.

\paragraph{\bf Ego-Exo Learning.}
Research on egocentric–exocentric video understanding primarily studies representation alignment and cross-view adaptation. For instance, prior efforts~\cite{sigurdsson2018charadesego, wang2023learning, xue2023learning, luo2025viewpoint} in action recognition have proposed self-supervised methods based on contrastive objectives for view-invariant representation learning. Another line of studies~\cite{li2021ego, luo2024put, zhang2025exo2ego} has explored distilling knowledge from one view to another. However, these efforts rarely examine whether the temporal reasoning of Video-LLMs remains consistent when identical events are observed from different viewpoints. In this work, we investigate this and propose a reinforcement learning approach to improve temporal reasoning consistency across viewpoints, inspired by the recent success of RL learning in video reasoning~\cite{feng2025videor1, wang2025time, liao2025improved, zhang2025tinyllava}.

\section{EgoExo-Con Dataset}
\subsection{Data Collection}
We source data from three datasets: CharadesEgo \cite{sigurdsson2018charadesego}, LEMMA~\cite{jia2020lemma}, and EgoExo-4D~\cite{grauman2024egoexo}, as they cover diverse and general domains, whereas other benchmarks are often restricted to specific domains (\eg, fitness \cite{li2024egoexofit}, toy assembly \cite{sener2022assembly101}, and surgery \cite{ozsoy2025egoexor}). CharadesEgo and LEMMA feature daily human-object interactions, while EgoExo-4D spans diverse skilled tasks, such as bike repairing and rock climbing. Among them, we collect synchronized video data annotated with query-timestamp pairs to support our focus on temporal understanding tasks. To ensure reliable evaluation, we carefully balance video diversity with model feasibility. For example, our preliminary experiments find that current models hardly perform effective temporal localization for long videos, making it infeasible to analyze consistency. Thus, we segment videos longer than five minutes into multiple clips surrounding the ground-truth moments, with each video clip lasting for at least two minutes, thus maximally preserving content diversity while keeping the task manageable for existing Video-LLMs.
\begin{figure}[t]
        \centering
        \includegraphics[width=1\linewidth]{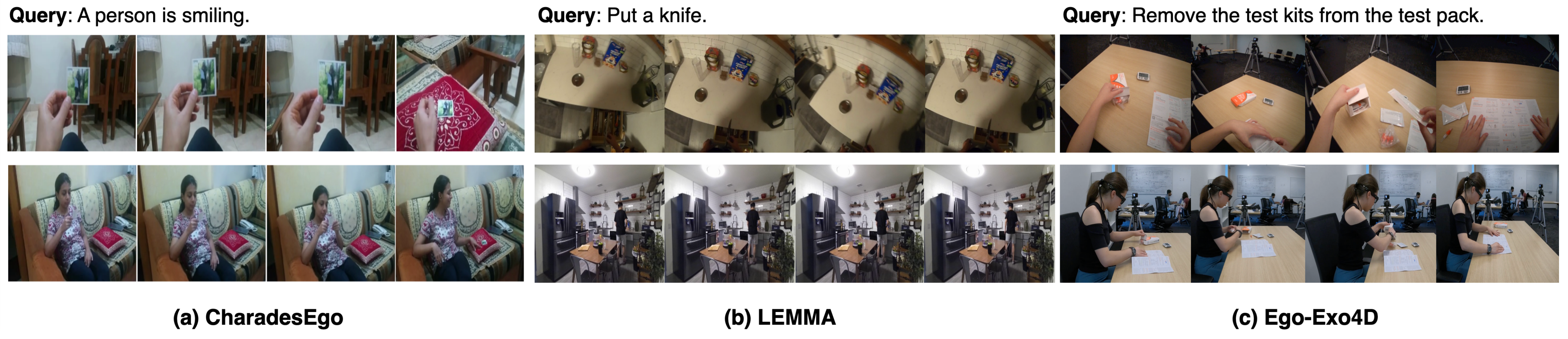}
        \vspace{-4mm}
    \caption{
    \textbf{Examples of queries and corresponding video moments from existing datasets.} (a) and (b) highlight fundamental limitations, with the egocentric view (top) in (a) being insufficient due to differing focuses, and the exocentric view (bottom) in (b) being ambiguous due to occlusion and distance. Although the query in (c) is identifiable from both viewpoints, we enrich it with details.
    }
    \label{fig: dataset_validation}
    \vspace{-5mm}
\end{figure}

\subsection{Data Filtering and Refinement}
Unfortunately, the original temporal queries in datasets often do not meet our requirements. In Fig.~\ref{fig: dataset_validation}, (a) queries in CharadesEgo are template-based, with categorical actions, and (b) queries in LEMMA rely on atomic actions and objects, both of which tend to miss details. More critically, viewpoint-induced ambiguities hinder reliable evaluation for cross-view consistency: key elements may be visible from one viewpoint but obscured from another due to varied focus and temporal alignment. For instance, the query ``A person is smiling'' in Fig.~\ref{fig: dataset_validation}-(a) is not visible from the egocentric video. In Appendix~\ref{appendix: unsuitable cases}, we also present unsuitable examples from a different dataset, EgoExo-Fitness. 

To address this issue, we reformulate queries in multiple stages. We convert the per-frame Human-Object Interaction (HOI) labels (\eg, put + cup, fridge) in LEMMA into natural-language queries. Since HOI labels are grounded to short 1–2 second intervals, we aggregate consecutive annotations into longer spans, extract salient verbs and nouns, and verbalize them into natural-language queries using simple rules for targets and prepositions. Next, we utilize a strong model (\ie, GPT-4o~\cite{achiam2023gpt4}) to enrich the original queries across all datasets. Specifically, given sampled frames from the target moments, the model verifies whether a query contains elements that cannot be reliably inferred from one or both viewpoints and produces refined alternatives. Additionally, the model generates a \emph{misaligned query} containing irrelevant content, which serves as a negative sample for temporal verification, thus balancing answers for ``yes'' and ``No'' in the verification task. The full prompt is shown in Appendix Fig.~\ref{fig: prompt_for_refinement}.

\begin{figure}[t]
        \centering
        \includegraphics[width=1\linewidth]{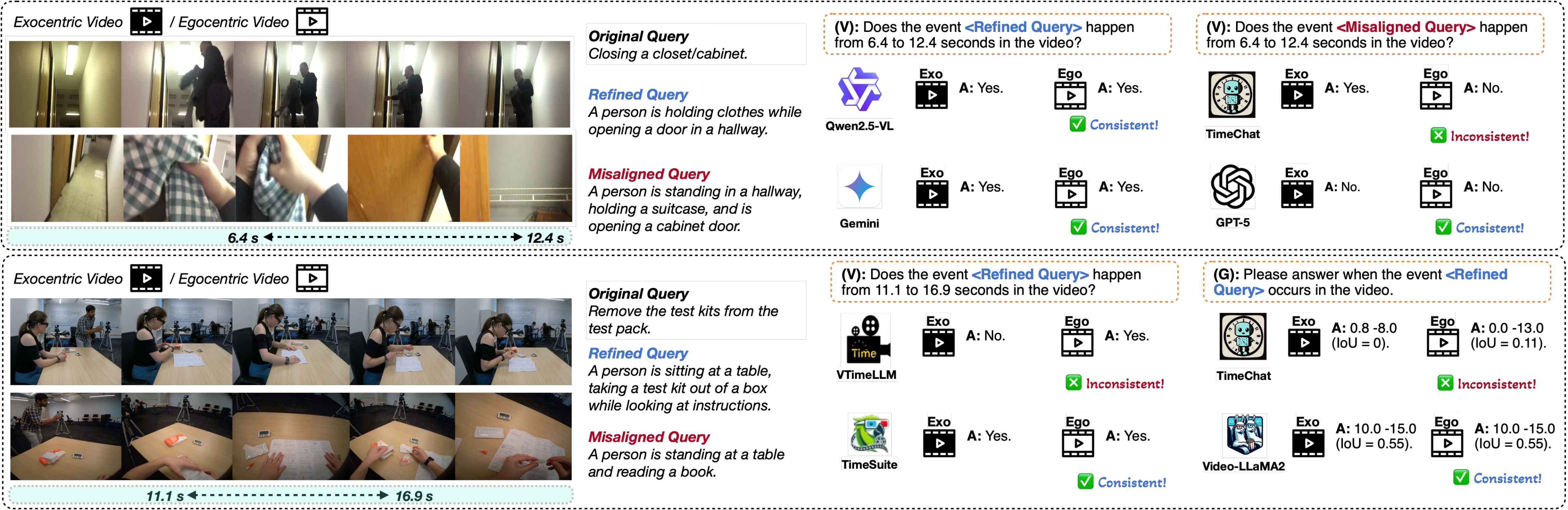}
    \caption{
    \textbf{Examples of test data and the corresponding model responses.} 
    We create refined and misaligned queries from each original query, use them for temporal verification (V) and grounding (G), and assess cross-view answer consistency.
    }
    \label{fig: visualization}
    \vspace{-5mm}
\end{figure}

Finally, we perform human validation for all samples. Four human evaluators review the generated queries alongside associated synchronized video pairs. They confirm whether refined queries are accurately grounded in both viewpoints, and misaligned queries intentionally conflict with the visual content. Queries passing this validation are retained, while ambiguous or low-quality samples are refined further or discarded (\eg, if the video itself is too noisy). Uncertain cases are cross-checked with the authors to ensure reliability. Fig.~\ref{fig: visualization} presents examples of refined and misaligned queries generated from the original query and its associated video.

\begin{figure}[t]
        \centering
        \includegraphics[width=1\linewidth]{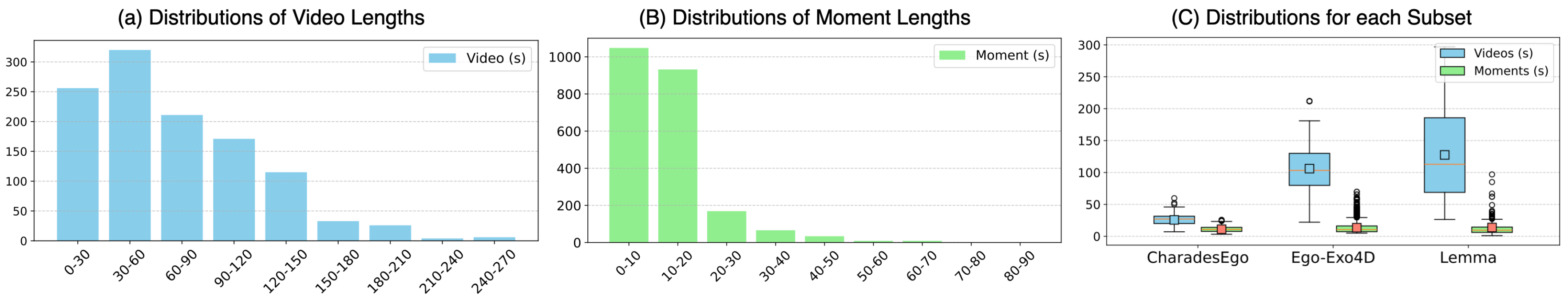}
        \vspace{-3mm}
    \caption{
    \textbf{Statistics of EgoExo-Con.} We show the distribution of video lengths (a), moment lengths (b), and each subset statistics (c). Each subset exhibits distinct temporal scales across videos and moments.
    }
    \label{fig: statistics}
\end{figure}

\subsection{Dataset Statistics}
\label{subsec: Statistics}
Eventually, we obtain 1,148 synchronized videos (426, 558, and 164 from CharadesEgo, EgoExo4D, and LEMMA, respectively) and 2,269 tightly aligned queries with timestamps. As shown in Fig.~\ref{fig: statistics}, each subset introduces distinct challenges specific to its domain and contributes a diverse range of video and moment lengths. Across the entire dataset, average duration and moment length are 86.4 seconds and 12.8 seconds, respectively. The average query length is 13.0 tokens, while the corresponding misaligned queries contain 15.9 tokens on average.

EgoExo-Con is a carefully curated video dataset that enables the evaluation of fine-grained, view-invariant temporal understanding, supported by extensive human verification and refinement. Among benchmarks that satisfy such strict criteria and involve human effort, EgoExo-Con remains competitive in scale. For instance, Video-MME~\cite{fu2024videomme} is a representative VideoQA benchmark consisting of 900 videos and 2,700 queries; however, those are not temporally annotated. EgoExo-Learn~\cite{huang2024egoexolearn} contains 747 videos with human demonstrations capturing processes across viewpoints, yet the videos are not synchronized.

\section{Evaluation}
\subsection{Models}
\paragraph{\bf Baseline.}
We evaluate a series of open-sourced models and categorize them as general-purpose or time-aware models, depending on whether they are designed for generic question-answering or specifically tuned to provide answers with corresponding video timestamps. Four general-purpose models: VideoChat2~\cite{li2024mvbench}, Qwen2.5-VL~\cite{bai2025qwen2}, Video-LLaMA2~\cite{cheng2024videollama2}, and Video-LLaMA3~\cite{zhang2025videollama3}, and four time-aware models: VTimeLLM~\cite{vtimellm}, TimeChat~\cite{timechat}, TimeSuite~\cite{zeng2024timesuite}, and TimeChat-VT~\cite{jung2025consistency}, are included. We provide details of each model in Appendix~\ref{appendix: model explanation}. Additionally, we include two powerful closed-source models: GPT-5~\cite{2025gpt5} and Gemini-2.5 Flash~\cite{comanici2025gemini}. We also benchmark human performance as a reference. We invite four evaluators and present each viewpoint independently to avoid biased predictions, reporting the average of their scores. Note that closed-source models and human performance are reported on a $\simeq$30\% subset of the full benchmark, which was uniformly sampled from each split to control evaluation costs. Additionally, we benchmark a random method that always returns ``yes'' or the entire video span for temporal verification and grounding, respectively.

\paragraph{\bf Evaluation Metric.}
We use accuracy in percentage for temporal verification (V) and R@1, Intersection over Union (IoU)=0.5 for temporal grounding (G). For grounding, predictions are considered correct if their IoU with the ground-truth moment exceeds 0.5. A model is evaluated separately for each viewpoint: V-Ego and V-Exo measure binary accuracy for egocentric and exocentric videos, respectively. Similarly, G-Ego and G-Exo denote grounding performance. Consistency metrics, V-EgoExo and G-EgoExo, measure whether a model correctly verifies or grounds specific moments (\ie, 0.5 $\le$ IoU) for both synchronized videos; consistent but wrong answers are not considered. 

\newcommand{\cmark}{\ding{51}}
\newcommand{\xmark}{\ding{55}}

\begin{table*}[t]
     \caption{\textbf{Existing model performance on EgoExo-Con.}  F: input frames. Ego: include ego data for training.
    }
    \vspace{-2mm}
    \small
    \resizebox{\linewidth}{!}{
      \begin{tabular}{lcc cccccc}
      \toprule
    \multirow{2}{*}{\bf Methods} & \multirow{2}{*}{\bf \# F} & \multirow{2}{*}{\bf Ego} & \multicolumn{6}{c}{\bf EgoExo-Con} \\
    \cmidrule(lr){4-9} 
       & & & V-Exo & V-Ego & V-ExoEgo & G-Exo & G-Ego & G-ExoEgo  \\ \midrule
        \rowcolor{lightgray} \cellcolor{white} Human & - & - & 92.1 & 91.3 & 89.4 & 72.4 & 73.0 & 67.3 \\  \midrule
        \textcolor{gray}{\textit{Closed-source}} \\ 
       \rowcolor{lightgray} \cellcolor{white} GPT-5~\cite{2025gpt5} & 32 & - & 60.5 & 61.3 & 52.5 & 34.5 & 32.8 & 20.1  \\
       \rowcolor{lightgray} \cellcolor{white} Gemini-2.5 Flash~\cite{comanici2025gemini} & 1 fps & - & 70.4& 70.1 & 52.3 & 42.0 & 45.9 & 20.8 \\ \midrule
       Random & - & - & 50.0 & 50.0 & 50.0 & 12.5 & 12.5 & 12.5 \\ \midrule
      \textcolor{gray}{\textit{General-purpose}} \\ 
      VideoChat2~\cite{li2024mvbench} & 16 & \cmark & 46.0 & 45.1 & 23.4 & 5.6 & 5.3 & 4.0 \\
        Qwen2.5-VL~\cite{bai2025qwen2} & 1 fps & \xmark & 54.3 & \underline{56.3} & 33.0 & 14.2 & 11.4 & 6.9  \\
        Video-LLaMA2~\cite{cheng2024videollama2} & 8 & \cmark & 53.3 & 52.1 & 27.9 & 12.0 & 11.5 & 7.5 \\
        Video-LLaMA3~\cite{zhang2025videollama3} & 1 fps & \cmark  & \underline{56.7} & 54.6 & \underline{36.6} & 27.7 & \bf 28.0 & 16.2 \\
        \midrule
        \textcolor{gray}{\textit{Time-aware}} \\
        VTimeLLM~\cite{vtimellm} & 100 & \xmark & 48.9 & 48.5 & 23.5 & 12.6 & 11.1 & 6.5 \\
        TimeChat~\cite{timechat} & 96 & \xmark & 48.9 & 48.4 & 25.1 & 21.3 & 20.5 & 12.8 \\
        TimeSuite~\cite{zeng2024timesuite} & 128 & \cmark & 47.4 & 48.5 & 25.6 & \bf \bf 28.2 & \underline{27.3} & \bf 18.7  \\ 
        TimeChat-VT~\cite{jung2025consistency} & 96 & \xmark & \bf 62.1 & \bf 61.4 & \bf 42.1 & \underline{27.8} & 26.2 & \underline{16.3} \\ 
      \bottomrule
      \end{tabular}
    }
    \label{tbl: main}
\end{table*}
\subsection{Performance Analysis on EgoExo-Con}
Our analyses on the results of EgoExo-Con in Table~\ref{tbl: main} are as follows: 
\textbf{(1) Single view vs. Cross view.} While all models show a modest performance gap between individual ego and exo videos, they struggle with cross-view consistency in both tasks. Open-source models in particular achieve barely half of their single-view performance. This demonstrates that the relatively stable performances across viewpoints are largely due to view-specific bias cues but not robust cross-view reasoning.
\textbf{(2) Time-aware vs. General-purpose.} Time‑aware models generally lead on grounding. TimeSuite attains the strongest grounding consistency, and TimeChat‑VT is competitive while also giving the best verification consistency. This advantage likely comes from better instruction tuning. However, VTimeLLM underperforms most general-purpose models, showcasing that time-aware models are not necessarily superior to general-purpose ones in temporal reasoning. 
\textbf{(3) Closed-sourced vs Human.} Closed-source models generally outperform open-source ones, reflecting their stronger capabilities. Yet a substantial gap (37\%~47\%) in cross-view consistency remains compared to humans, with scores approaching random, underscoring the challenge of EgoExo-Con and the room for improvement. 
\textbf{(4) Training with ego data is not sufficient.} Models including egocentric videos for training do not consistently yield higher consistency than others trained on exocentric videos alone, showing that simply mixing ego and exo data does not benefit consistency. 
\textbf{(5) Temporal reasoning outweighs increasing frames for better results.} Video‑LLaMA2 (8 frames) outperforms VideoChat2 (16 frames) across all metrics. TimeChat‑VT (96 frames) also outperforms several models that use more/less context, suggesting reasoning and temporal modeling matter more than sheer frame count. 

\begin{figure}[t]
        \centering
        \includegraphics[width=1\linewidth]{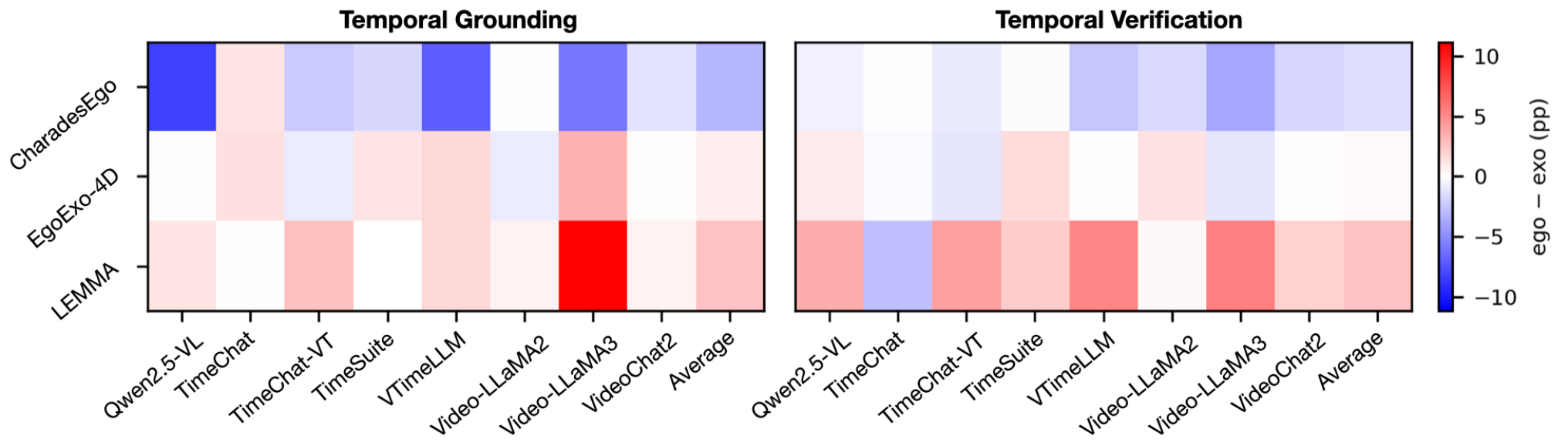}
    \caption{
    \textbf{Heatmaps of the performance gap.} All values are reported in percentage points. Red and blue indicate higher performances on \emph{ego} and \emph{exo} perspectives, respectively. \textit{\ie, a blue cell indicates that the corresponding model performs better on exocentric views than on egocentric ones, while red indicates better performance on egocentric views compared to exocentric ones.}
    }
    \label{fig: heatmap across subsets and models}
    \vspace{-2mm}
\end{figure}

Furthermore, we analyze model behaviors across subsets by plotting the ego–exo performance gap in Fig.~\ref{fig: heatmap across subsets and models}. Patterns are consistent across grounding and verification: in CharadesEgo, most models perform better on exocentric views (blue), whereas in LEMMA, they tend to favor egocentric views (red); EgoExo-4D shows mixed but generally smaller gaps. These trends likely reflect domain characteristics. As shown in Fig.~\ref{fig: visualization}, when a person performs in a fixed position, egocentric videos often provide favorable cues such as clearer hand–object interactions. Conversely, when a person moves around or changes location, egocentric views can become more challenging due to rapid scene shifts, while exocentric views offer greater stability. Such trends are particularly pronounced in CharadesEgo. Detailed results across subsets are provided in Appendix Table~\ref{tbl: performance across subsets}. Although the relative effectiveness of each viewpoint varies by domain, they do not significantly affect overall consistency. Overall, across models, consistency scores lag far behind single-view metrics, underscoring that achieving robust, view-invariant temporal understanding remains an open challenge.

\subsection{Supervised Performance on EgoExo-Con}
\label{subsec: supervised}
In this section, we study whether supervised fine-tuning (SFT) with synchronized ego-exo video data improves performance.
We first collect 3.6k from CharadesEgo and 2.3k videos from EgoExo-4D training sets. We exclude LEMMA due to its limited size (\ie, 243 videos for training). Then we finetune two general-purpose models: VideoChat2 and Video-LLaMA2 for temporal verification, and two time-aware models: TimeChat and TimeSuite for temporal grounding. All models apply  LoRA~\cite{hu2022lora} fine-tuning with their official configurations. More implementation details are presented in Appendix~\ref{appendix: Details of Supervised Fine-tuning for Video-LLMs}.

\begin{table*}[t]
    \caption{
    \textbf{Fine-tuned performance.} The left and right tables report model performance for temporal verification (V) and temporal grounding (G). The first row in each model reports zero-shot performance, while the subsequent rows present results after fine-tuning on either (+Ego, +Exo) or both viewpoints (+ EgoExo). Notably, models trained on both viewpoints are not always the best-performing, and then they barely outperform those trained on a single viewpoint.
    }
    \vspace{-2mm}
    \centering
    \begin{minipage}{0.5\linewidth}
    \resizebox{\linewidth}{!}{
    \begin{tabular}{l ccc ccc}
    \toprule
    \bf \multirow{2}{*}{Methods} & \multicolumn{3}{c}{\bf CharadesEgo} & \multicolumn{3}{c}{\bf EgoExo-4D} \\ 
    \cmidrule(lr){2-4} \cmidrule(lr){5-7}
    & V-Exo & V-Ego & V-ExoEgo & V-Exo & V-Ego & V-ExoEgo \\ \midrule
    \rowcolor{lightgray} \cellcolor{white} VideoChat2 & 46.3 & 44.4 & 22.2 & 41.4 & 41.5 & 19.8 \\ 
    + Ego & 56.2 & \bf 59.2 & \bf 36.4 & 46.5 & 49.2 & 29.7 \\
    + Exo & \bf 56.6 & 57.5 & 35.5 & 46.1 & 47.3 & 29.4 \\
    + EgoExo & 56.4 & 57.1 & 34.7 & \bf 48.3 & \bf 50.1 & \bf 30.1 \\
         \midrule
         \rowcolor{lightgray} \cellcolor{white} Video-LLaMA2 & 54.0 & 52.4 & 28.2 & 51.5 & 52.8 & 28.1 \\ 
         + Ego & 57.0 & \bf 58.2 & \bf 31.7 & \bf 60.2 & \bf 60.3 & 38.8 \\
         + Exo & 57.6 & 56.1 & 31.4 & 59.8 & 60.1 & \bf 39.2 \\
         + EgoExo & \bf 58.5 & 57.3 & 31.0 & 57.5 & 59.6 & 39.1 \\
      \bottomrule
      \end{tabular}
      }
      \vfill
    \end{minipage} 
    \begin{minipage}{0.48\linewidth}
    \resizebox{\linewidth}{!}{
    \begin{tabular}{l ccc ccc}
    \toprule
    \bf \multirow{2}{*}{Methods} & \multicolumn{3}{c}{\bf CharadesEgo} & \multicolumn{3}{c}{\bf EgoExo-4D} \\ 
    \cmidrule(lr){2-4} \cmidrule(lr){5-7}
    & G-Exo & G-Ego & G-ExoEgo & G-Exo & G-Ego & G-ExoEgo \\ \midrule
        \rowcolor{lightgray} \cellcolor{white} TimeChat &  44.9 & 46.1 & 30.1 & 4.9 & 6.3 & 3.3 \\ 
        + Ego & \bf 62.0 & \bf 62.1 & \bf 48.3 & 9.7 & \bf 13.1 & \bf 4.9 \\
        + Exo & 58.8 & 60.0 & 47.1 & 10.4 & 10.7 & 4.2 \\
        + EgoExo & 60.3 & 61.8 & 46.5 & \bf 10.6 & 11.5 & 4.3 \\ 
         \midrule
         \rowcolor{lightgray} \cellcolor{white} TimeSuite & 63.4 & 56.2 & 44.8 &  5.8 & 8.3 & 2.3 \\ 
         + Ego & 61.0 & 61.5 & 54.0 & 6.7 & 6.2 & 4.8 \\
         + Exo & \bf 74.6 & \bf 68.7 & \bf 59.5 & \bf 10.5 & \bf 9.9 & \bf 5.7 \\
         + EgoExo & 67.8 & 61.1 & 51.3 & 9.6 & 8.9 & 5.3 \\ 
      \bottomrule
      \end{tabular}
      }
      \vfill
    \end{minipage}
    \label{tbl: supervised}
    \vspace{-3mm}
\end{table*}
Table~\ref{tbl: supervised} reports results of three different settings: training on either viewpoint (Ego, Exo) or both viewpoints together (ExoEgo). All of them consistently improve over zero-shot baselines. Interestingly, despite utilizing twice the data, training on both viewpoints (ExoEgo) yields only marginal gains and often underperforms the models trained on a single view. Specifically, in CharadesEgo, TimeSuite shows a notable 8.1\% gap in consistency between training on both viewpoints and training only on exocentric videos. Furthermore, the improvements remain limited even when unfreezing the visual encoder (refer to Appendix Table~\ref{tbl: unfreeze}), suggesting that simply increasing model capacity is insufficient and highlighting the fundamental challenge of view-invariant temporal understanding. 

The above findings are consistent with observations in cross-view learning~\cite{li2024egoexofit, sigurdsson2018charadesego}, where naively blending perspectives does not always bring improvement. Without explicit alignment, conflicting priors across tasks or domains undermine temporal signals and consistency rather than improvement.
\begin{figure}[t]
        \centering
        \includegraphics[width=1\linewidth]{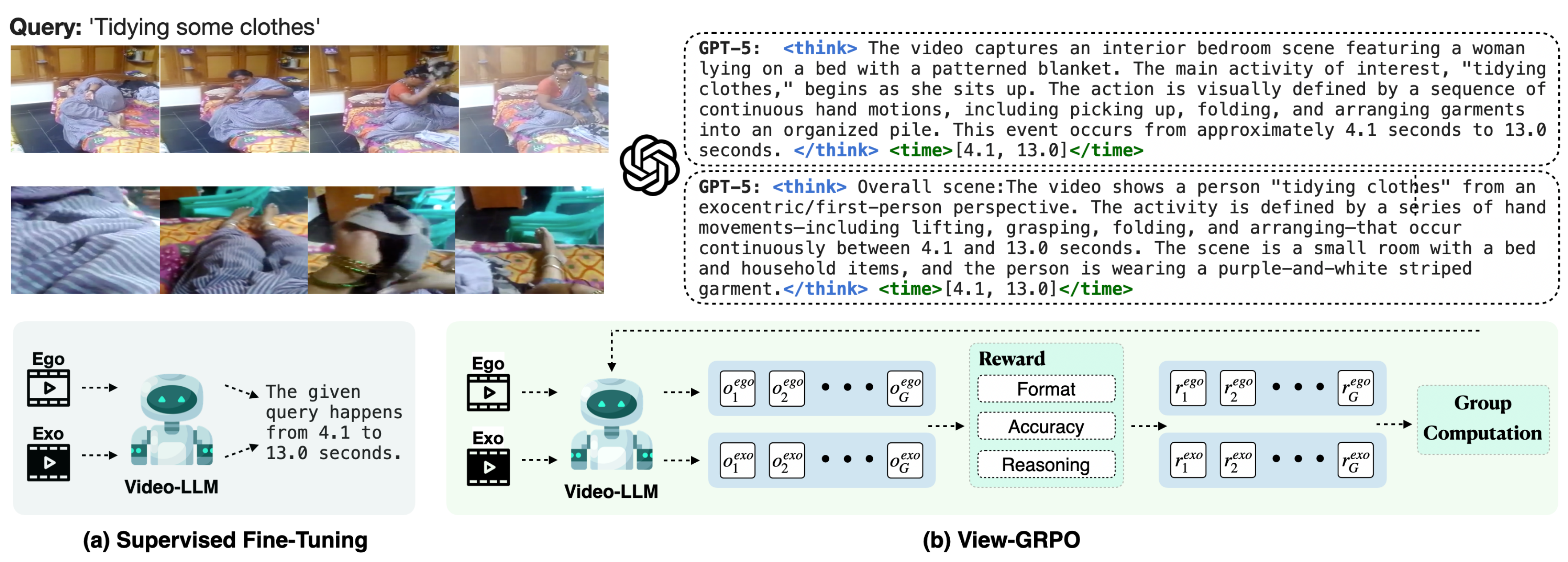}
    \caption{
    \textbf{Overview of our approach.} 
    (a) In supervised fine-tuning, the model is trained to directly predict the same query answers (\eg, video moments) for synchronized video pairs. (b) View-GRPO trains a model to provide viewpoint-specific reasoning chains, which are generated by GPT-5 (top).
    }
    \label{fig: method}
    \vspace{-5mm}
\end{figure}

\section{Method}
We propose a reinforcement learning (RL) framework that guides models toward developing viewpoint-specific reasoning while encouraging shared consistency. Rather than simply enforcing identical outputs, our approach encourages the model to internalize the robust reasoning traces with structured guidance, while simultaneously preserving agreement between the outputs for each viewpoint. We build on Group Relative Policy Optimization (GRPO), which is particularly well-suited as it leverages relative rewards instead of absolute scores.

\subsection{Background: Group Relative Policy Optimization (GRPO)}
GRPO~\cite{shao2024deepseekmath} is a reinforcement learning algorithm designed to refine large language model outputs by leveraging relative ranking among multiple candidate responses. Instead of treating each response independently with absolute rewards, GRPO evaluates a set of responses produced for the same prompt, assigning rewards in relation to the group. This group-wise normalization helps reduce reward variance and makes optimization more stable compared to approaches relying solely on absolute scores or pairwise comparisons.

Given a prompt $p$, the model generates $G$ candidate responses $o = \{o_1, \ldots, o_G\}$. Each response receives a reward value $r(o_i)$. GRPO then standardizes these scores within the group and optimizes a weighted objective:
\begin{equation}
    R(o) = \sum_{i=1}^{G} \frac{\pi_\theta(o_i)}{\pi_{\theta_{\text{old}}}(o_i)} 
    \cdot \frac{r(o_i) - \text{mean}(\{r(o_i)\}_{i=1}^G)}{\text{std}(\{r(o_i)\}_{i=1}^G)} ,
\end{equation}
where $\pi_\theta(o)$ denotes the current policy and $\pi_{\theta_{\text{old}}}(o)$ is the previous policy. To prevent divergence from the base model, KL-divergence regularization is added:
\begin{equation}
    \max_{\pi_\theta} \;
    \mathbb{E}_{o \sim \pi_{\theta_{\text{old}}}(p)}
    \Big[ R(o) - \beta D_{\text{KL}} \big( \pi_\theta \; \| \; \pi_{\text{ref}} \big) \Big],
\end{equation}
where $\pi_{ref}$ is the base model and $\beta$ controls the strength of the regularization. Please refer to the original paper for more details.

\subsection{Learning Temporal Reasoning across Viewpoints}
To adapt GRPO for cross-view reasoning, we first curate training data, including temporal reasoning chains for both egocentric and exocentric views. Fig.~\ref{fig: method} (Top) illustrates the generation of video reasoning data. We prompt GPT-5 to produce step-by-step reasoning chains for each video in order to solve the given task, ensuring viewpoint-specific reasoning while aligning the final answers. We discard samples lacking valid reasoning in either view, as these typically indicate ambiguous or low-quality data. Specifically, we exclude cases where the model explicitly states its failure in the answer, or where the predicted moment has a temporal IoU (tIoU) below 0.7 with the ground-truth. After filtering, we retain 3.3k videos with 30k reasoning instances, which we name as View30K.

We then design reward functions comprising three major components:
\paragraph{1. Format Reward.} For structured reasoning and easy answer extraction, responses must follow the template:
\texttt{<think>...</think><answer>...</answer>}. Formally:
\begin{equation}
    r_{\text{format}}(o) =
    \begin{cases}
        1, & \text{if $o$ follows the required format}, \\
        0, & \text{otherwise}.
    \end{cases}
\end{equation}

\paragraph{2. Accuracy Reward.} We unify task-specific accuracy into $r_{\text{acc}}$. For temporal grounding, the reward is the temporal Intersection-over-Union (tIoU) between ground truth $[t_s, t_e]$ and prediction $[t'_s, t'_e]$. For verification, it is binary correctness:
\begin{equation}
    r_{\text{accuracy}}(o) =
    \begin{cases}
        \tfrac{|[t_s, t_e] \cap [t'_s, t'_e]|}{|[t_s, t_e] \cup [t'_s, t'_e]|}, & \text{if \textit{grounding}}, \\[10pt]
        \mathbbm{1}[o = o^*], & \text{if \textit{verification}}.
    \end{cases}
\end{equation}

\paragraph{3. Reasoning Reward.} We design a unified reasoning reward that encourages both semantic alignment with reference reasoning $o^*$ and cross-view structural consistency with its counterpart reasoning $\tilde{o}$. Specifically, for a given video instance with two viewpoints, we denote by $o$ the reasoning generated from one viewpoint, and by $\tilde{o}$ the reasoning generated from its counterpart viewpoint. Without loss of generality, if $o=o^{ego}$, then $\tilde{o}=o^{exo}$, and vice versa. First, given a generated reasoning $o$ and a reference reasoning $o^*$, we employ an LLM judge to compute a semantic reward:
    \begin{equation}
    r_{\text{sem}}(o) = \mathrm{Judge}(o, o^*) \in [0,1].
    \end{equation}
This term encourages the generated reasoning to remain faithful to the reference explanation. Consequently, we extract action and object sets from $o$ and $\tilde{o}$ using spaCy, denoted as $\mathcal{A}$ and $\mathcal{O}$, respectively. We then measure their structural consistency via the Jaccard overlap: 
\begin{equation} 
    r_{\text{struct}} (o) = \frac{1}{2}\big(
    \text{Jac}(\mathcal{A}(o), \mathcal{A}(\tilde{o})) + 
    \text{Jac}(\mathcal{O}(o), \mathcal{O}(\tilde{o}))
    \big).
\end{equation}
The final reasoning reward integrates both components:
\begin{equation}
r_{\text{reasoning}} (o) =
    \lambda * r_{\text{sem}}(o) + (1 - \lambda) * r_{\text{struct}}(o),
\end{equation}
where $\lambda \in [0,1]$ balances reference alignment and cross-view consistency.

Overall, the total reward integrates all the components to balance correctness and reasoning quality:
\begin{equation}
    r(o) = r_{\text{accuracy}}(o) + r_{\text{format}}(o) + r_{\text{reasoning}}(o),
\end{equation}
thus enabling models to learn correct and consistent cross-view temporal reasoning. We name the overall approach View-GRPO, as shown in Fig.~\ref{fig: method}-(b) (Bottom).

\begin{table*}[t]
    \centering
    \small
    \caption{
    \textbf{Performance of View-GRPO on EgoExo-Con.}
    }
    \vspace{-2mm}
    \begin{tabular}{l cccccc}
    \toprule
    \bf \multirow{2}{*}{Methods} & \multicolumn{6}{c}{\bf EgoExo-Con} \\ 
    \cmidrule(lr){2-7}
   & V-Exo & V-Ego & V-ExoEgo & G-Exo & G-Ego & G-ExoEgo \\ \midrule
     \rowcolor{lightgray} Qwen2.5-VL-3B \cellcolor{white} & 51.0 & 52.5 & 28.1 & 10.1 & 9.9 & 7.9 \\
     + SFT & \underline{52.7} & 51.2 & \underline{30.6} & \underline{16.3} & \underline{16.6} & \underline{12.9} \\
     + GRPO & 52.5 & \underline{52.9} & 30.3 & 15.4 & 16.3 & 10.3 \\     
     + \textbf{View-GRPO} 
     & \bf 54.6\color{green}{ $\uparrow$3.6} 
     & \bf 54.4\footnotesize{\color{green}{ $\uparrow$1.9}} 
     & \bf 34.2\footnotesize{\color{green}{ $\uparrow$6.1}} 
     & \bf 18.9\footnotesize{\color{green}{ $\uparrow$8.8}} 
     & \bf 17.9\footnotesize{\color{green}{ $\uparrow$8.0}} 
     & \bf 14.9\footnotesize{\color{green}{ $\uparrow$7.0}} 
     \\ \midrule
     \rowcolor{lightgray} Qwen2.5-VL-7B \cellcolor{white} & 54.3 & 56.3 & 33.0 & 14.2 & 11.4 & 6.9  \\
     + SFT & \underline{57.6} & \underline{58.0} & \underline{41.4} & 18.3 & \underline{17.8} & \underline{14.9} \\
     + GRPO & 55.2 & 57.6 & 39.8 & \underline{18.6} & 16.1 & 14.3 \\
     + \textbf{View-GRPO} 
    & \bf 58.6\footnotesize{\color{green}{ $\uparrow$4.3}} 
     & \bf 58.2\footnotesize{\color{green}{ $\uparrow$1.9}}
     & \bf 45.1\footnotesize{\color{green}{ $\uparrow$12.1}} 
     & \bf 22.0\footnotesize{\color{green}{ $\uparrow$7.8}} 
     & \bf 21.6\footnotesize{\color{green}{ $\uparrow$10.2}} 
     & \bf 18.7\footnotesize{\color{green}{ $\uparrow$11.8}} 
     \\ \midrule
    \rowcolor{lightgray} \cellcolor{white} InternVL3.5-8B & 64.4 & 64.7& 50.7 & 12.8 & 6.7 & 3.0 \\
    + \bf View-GRPO 
    & \bf 73.1\footnotesize{\color{green}{ $\uparrow$8.7}} 
    & \bf 74.4\footnotesize{\color{green}{ $\uparrow$9.7}} 
    & \bf 62.4\footnotesize{\color{green}{ $\uparrow$11.7}} 
    & \bf 20.5\footnotesize{\color{green}{ $\uparrow$7.7}} 
    & \bf 16.8\footnotesize{\color{green}{ $\uparrow$10.1}} 
    & \bf 10.6\footnotesize{\color{green}{ $\uparrow$7.6}} \\ 
    \bottomrule
      \end{tabular}
    \label{tbl: method}
\end{table*}
\subsection{Implementations}
We use Qwen2.5-VL~\cite{bai2025qwen2} and InternVL3.5~\cite{wang2025internvl3} as a backbone. During training, all experiments set the same frame sampling rate (\ie, 2 FPS) and freeze the visual encoder and update only the parameters of the LLM. The training involves 8 $\times$ A100 GPUs, and further implementation details are in Appendix~\ref{appendix: details of View-GRPO}.

\subsection{Analyses on View-GRPO}
Table~\ref{tbl: method} shows the performance of our method, View-GRPO, on EgoExo-Con compared to standard SFT and GRPO implementations. View-GRPO demonstrates superior performance, achieving significant improvements over baselines beyond those obtained with SFT and GRPO. Notably, the most significant improvements are often on cross-view consistency (V-ExoEgo and G-ExoEgo), although it also benefits individual views. We conjecture that the reasoning reward plays a central role, as it delivers noticeably higher consistency compared to naive GRPO. Additionally, View-GRPO remains effective with the different backbone, InternVL3.5. By encouraging models to produce faithful, step-by-step temporal explanations tailored to each viewpoint while converging toward consistent temporal conclusions, the model reduces reliance on view-specific biases and instead learns shared temporal abstractions.
\begin{table}[t]
    \centering
    \begin{minipage}[t]{0.54\textwidth}
        \caption{\textbf{Ablation on reasoning reward.} 
        }
        \vspace{2mm}
        \centering
        \resizebox{\linewidth}{!}{
          \begin{tabular}{cc cccccc}
          \toprule
        \multirow{2}{*}{\bf $r_{\text{sem}}$} & \multirow{2}{*}{\bf $r_{\text{struct}}$} & \multicolumn{6}{c}{\bf EgoExo-Con} \\
        \cmidrule(lr){3-8} 
           & & V-Exo & V-Ego & V-ExoEgo & G-Exo & G-Ego & G-ExoEgo  \\ \midrule
            \rowcolor{lightgray} \cellcolor{white} & \cellcolor{white} & 55.2 & 57.6 & 39.8 & 18.6 & 16.1 & 14.3 \\
            \checkmark & & 58.3 & 58.1 &  44.7  & 21.5  & 21.0 & 18.3 \\
            \checkmark & \checkmark & \bf 58.6 & \bf 58.2 & \bf 45.1 & \bf 22.0 & \bf 21.6 & \bf 18.7 \\
          \bottomrule
          \end{tabular}
        }
        \label{tbl: reasoning reward}
    \end{minipage}
    \hfill
    \begin{minipage}[t]{0.43 \textwidth}
         \caption{\textbf{Performance on Video-MME and TVGBench.}}
         \vspace{-2mm}
    \resizebox{\linewidth}{!}{
          \begin{tabular}{l c ccc}
          \toprule
        \multirow{2}{*}{\bf Methods} & \bf Video-MME & \multicolumn{3}{c}{\bf TVGBench} \\
        \cmidrule(lr){2-2} \cmidrule(lr){3-5} 
           & w/o subs & R1@0.3 & R1@0.5 & R1@0.7  \\ \midrule
           TimeChat & 30.2 & 22.4 & 11.9 & 5.3 \\
           TimeSuite & 46.3 & \underline{31.1} & 18.0 & 8.9 \\
           Qwen2.5-VL & \underline{61.1} & 28.1 & \underline{19.5} & \underline{10.5} \\
           \bf View-GRPO & \bf 69.7 & \bf 42.0 & \bf 25.0 & \bf 13.9 \\
          \bottomrule
          \end{tabular}
          }
           \label{tbl: otherbm}
    \end{minipage}
    \vspace{-6mm}
\end{table}

\begin{wrapfigure}[12]{r}{0.5\textwidth}
    \centering
    \includegraphics[width=1.0\linewidth]{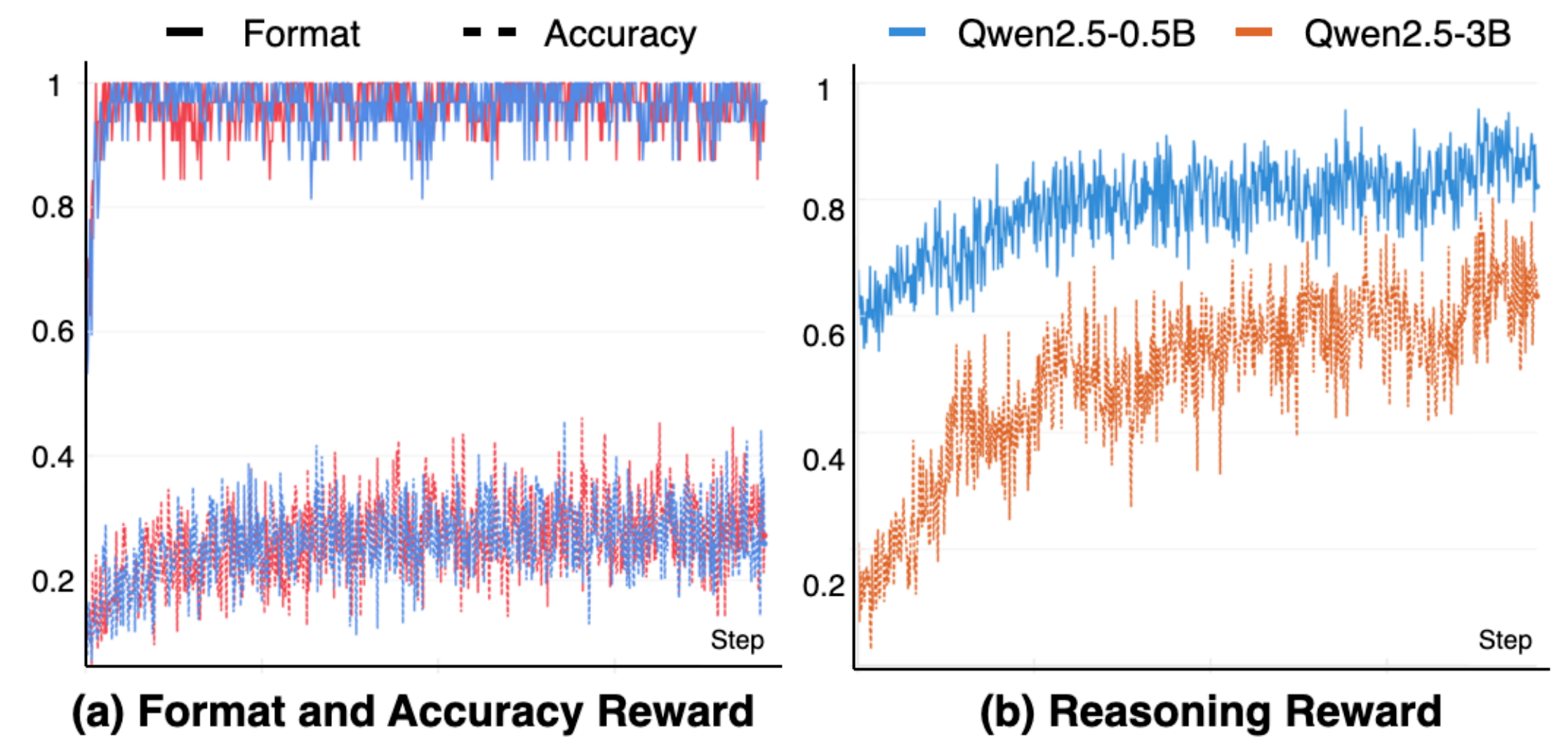}
    \caption{
   \textbf{Reward of different LLM judges.} Qwen2.5-0.5B raises calibration concerns due to its overly high reasoning rewards from early steps.
    }
    \label{fig: ablation}
\end{wrapfigure}
We further analyze our reasoning reward design in depth. While LLMs are commonly used as judges~\cite{zheng2023judging, xie2023text2reward}, the role in optimizing models in View-GRPO remains underexplored despite its effectiveness, as they often introduce potential bias and uncertainty in video evaluation~\cite{cores2024lost, liu2025your}. To provide insights into this, we employ judge models of different scales for temporal grounding and analyze their impact on optimization. In Fig.~\ref{fig: ablation}-(a), format and accuracy rewards remain relatively stable across scales. However, Qwen2.5-0.5B produces overly high reasoning rewards from the very first training steps in Fig.~\ref{fig: ablation}-(b), raising concerns about calibration and reliability. We find that this behavior leads to a measurable consistency degradation (\ie, –3\% in G-EgoExo). Table~\ref{tbl: reasoning reward} shows the effect of different components in $r_{\text{reasoning}}$. The semantic reward $r_{\text{sem}}$ consistently improves performance over the baseline, and incorporating the structural consistency $r_{\text{struct}}$ leads to further gains. By $r_{\text{sem}}$ for conceptual alignment with reference reasoning and $r_{\text{struct}}$ for cross-view alignment, the model achieves more consistent temporal reasoning across viewpoints. We further investigate performance across varying reasoning lengths to assess the potential impact of hallucinations in reasoning in Appendix~\ref{appendix: reasoning length}. Overall, such results underscore the critical influence of LLM-judges and reward calibration on model optimization.

Additionally, we evaluate our method on additional benchmarks: Video-MME~\cite{fu2024videomme} and TVGBench~\cite{wang2025time}. We select Video-MME as a representative benchmark for VideoQA, while TVGBench encompasses diverse temporal grounding datasets with highly varied query formulations. Table~\ref{tbl: otherbm} demonstrates that our method possesses strong generalization capabilities across diverse video benchmarks. We attribute this robustness to strengthened temporal reasoning and improved cross-view consistency through View-GRPO. We discuss potential directions for improvement from several perspectives: data, architecture, and learning, in Appendix~\ref{appendix: discussion}. Overall, our method demonstrates its effectiveness and potential for consistent and view-invariant temporal understanding.

\begin{figure}[t]
        \centering
        \includegraphics[width=1\linewidth]{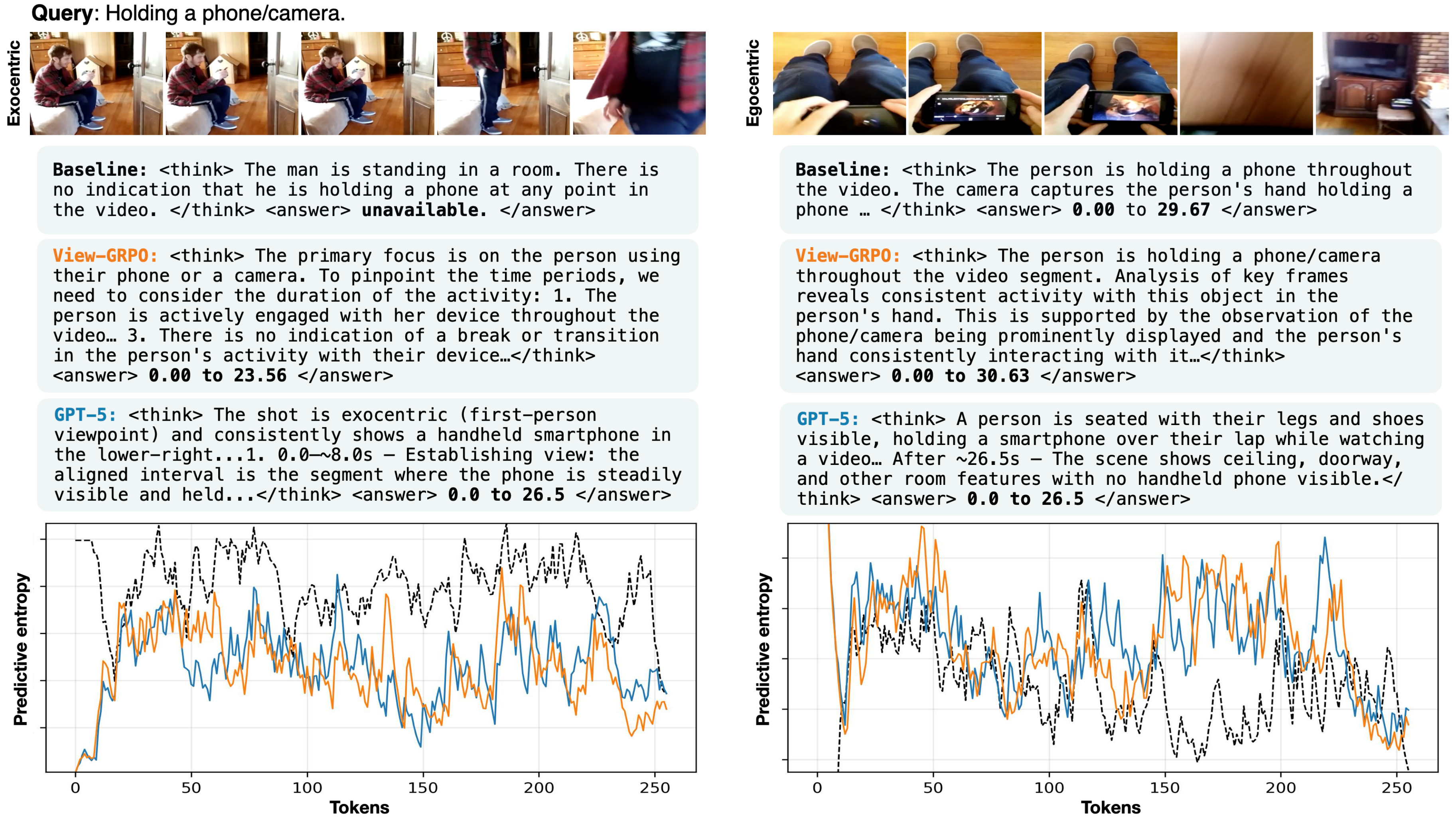}
    \caption{
    \textbf{Visualization of View-GRPO.} We visualize the model-generated reasoning and the reference reasoning for each viewpoint, along with their token-level entropy distributions over the generation length. The model produces step-by-step temporal reasoning with accurate grounding predictions, where the high overlap in token entropy patterns indicates close alignment with GPT-5’s reasoning traces.
    }
    \label{fig: View-GRPO visualization}
\end{figure}

\subsection{Qualitative Analysis}
Fig.~\ref{fig: View-GRPO visualization} visualizes a qualitative comparison of reasoning traces and token-level predictive entropy across viewpoints. We compare the baseline model (\ie, Qwen2.5-VL-7B), GPT-5 for reference, and our method under synchronized video pairs. Although the baseline provides an accurate answer for the egocentric video, it fails to identify the target moment for the exocentric video, showing unreliable reasoning behavior and inconsistency. In contrast, View-GRPO provides the step-by-step temporal reasoning traces with accurate grounding and closely follows the reference reasoning, exhibiting similar peaks and transitions. This suggests that View-GRPO internalizes robust reasoning patterns while aligning its consistent answers across viewpoints. 
\section{Conclusion}
In this work, we introduced EgoExo-Con that comprises synchronized egocentric and exocentric videos paired with human-refined queries and evaluated whether models perform consistent temporal understanding across viewpoints. EgoExo-Con revealed that Video-LLMs struggle with cross-view consistency, lagging behind single-viewpoint performance, and that naively training on both viewpoints does not reliably help. To address this, we proposed View-GRPO that encourages viewpoint-specific temporal reasoning while promoting cross-view alignment, demonstrating its effectiveness over alternative training strategies. We hope EgoExo-Con and View-GRPO pave the way toward truly robust and view-invariant temporal understanding.


\clearpage  


%
%
\bibliographystyle{splncs04}
\bibliography{main}

\appendix
\clearpage
\setcounter{section}{0}
\appendix
\section*{Appendix}
We provide further details that are not included in the main paper due to the page limitation.


    


\section{Details of Models}
\label{appendix: model explanation}
In this section, we describe eight open-source \vllm: VideoChat2~\cite{li2024mvbench}, Qwen2.5-VL~\cite{bai2025qwen2}, Video-LLaMA2~\cite{cheng2024videollama2}, Video-LLaMA3~\cite{zhang2025videollama3}, VTimeLLM~\cite{vtimellm},  TimeChat~\cite{timechat}, TimeSuite~\cite{zeng2024timesuite}, and TimeChat-VT~\cite{jung2025consistency}, which are utilized in our evaluation. Note that the size of the models is 7B.

\begin{enumerate}[leftmargin=*]
\item \textbf{VideoChat2} design three different video instruction tuning stages. Specifically, they align multimodal inputs in the first stage and then generate captions from various image-text pairs. Finally, they conduct instruction tuning to better align responses with human instructions. VideoChat2 demonstrates significant improvements in video question answering benchmarks in zero-shot settings.

\item \textbf{Qwen2.5-VL} Qwen2.5-VL is one of the latest models of Qwen vision-language series. It achieves strong benchmark performances through enhanced visual recognition, precise object localization, robust document parsing, and long-video comprehension.

\item \textbf{Video-LLaMA2} is one of the state-of-the-art \vllm, demonstrating superior performances on video question-answering tasks. They seamlessly integrate both visual and audio modalities in videos and propose STC connector to understand spatiotemporal video information. 

\item \textbf{Video-LLaMA3} is a series of Video-LLaMA family. It emphasizes a vision-centric training paradigm and vision-centric framework design. With curated high-quality image and video data, Video-LLaMA3 achieves compelling performances across diverse image and video understanding benchmarks.

\item \textbf{VTimeLLM} aims to include video timestamps along with the answer for human instruction. It designs a three-stage instruction tuning. Initially, it trains a visual projection layer with image-text pairs and then incorporates video datasets with temporal modeling. VTimeLLM devises two types of QA dialogue templates, including single-turn and multi-turn, to prompt questions requiring a comprehensive description of all events and their corresponding timestamps. 

\item \textbf{TimeChat} is developed to localize and identify specific video moments from a given human instruction. It utilizes a time-aware frame encoder that injects timestamp information into visual features, leveraging Q-Former, and designs a sliding video Q-Former to handle temporal information. 

\item \textbf{TimeSuite} tackles achieving high performance on both question-answering and grounding for long videos. It argues that previous \vllm~struggle to achieve both capabilities and develop a VideoChat-T and a temporal-centric instruction-tuning dataset, TimePro.

\item \textbf{TimeChat-VT} is a model that specifically takes into account consistency modeling. Upon TimeChat, it develops a new instruction tuning method, VTune, that converts temporal grounding into a verification process. This requires not only precise temporal grounding, but also confirming the occurring events from specific video moments.
\end{enumerate}

\section{Details of Experiments}
\subsection{Prompt Templates}
For temporal verification, we give variations to prompt templates beyond utilizing misaligned queries. Specifically, following the previous work~\cite{jung2025consistency}, we also include templates like ``Is the event missing in the video?'' or ``Is the event not present in the video?'' to shift the correct answer ``Yes.'' to ``No.''

For temporal grounding, some general-purpose models do not officially provide prompt templates for grounding. Therefore, we borrow the prompts from the previous work~\cite{jung2025consistency} for the general-purpose models and closed-sourced models, and design the prompt \textit{``Give the query, when does the described content occur in the video? Please return its start and end time using `start - end seconds'.''} to ensure the model includes timestamps in its answer.

\subsection{Details of Supervised Fine-tuning for Video-LLMs}
\label{appendix: Details of Supervised Fine-tuning for Video-LLMs}
For each task, we design question–answer templates. In temporal verification, we use two formats: ``Does `event' happen from ‘st’ to ‘ed’ in the video?'' and ``Does `event' not happen from ‘st’ to ‘ed’ in the video?''. Here, `event', `st', and `ed' are replaced with the annotated query and its start and end timestamps. Answers are restricted to ``yes'' or ``no.'' For temporal grounding, the template ``Localize the `event' in the video and return its start and end times.'' is employed. Note that both tasks utilize the same number of videos and queries. Again, we follow the official code and configurations in each model. All experiments run 3 epochs with 4 $\times$ A100 GPUs.

\subsection{Details of View-GRPO training}
\label{appendix: details of View-GRPO}
\begin{figure}[t]
        \centering
        \includegraphics[width=1\linewidth]{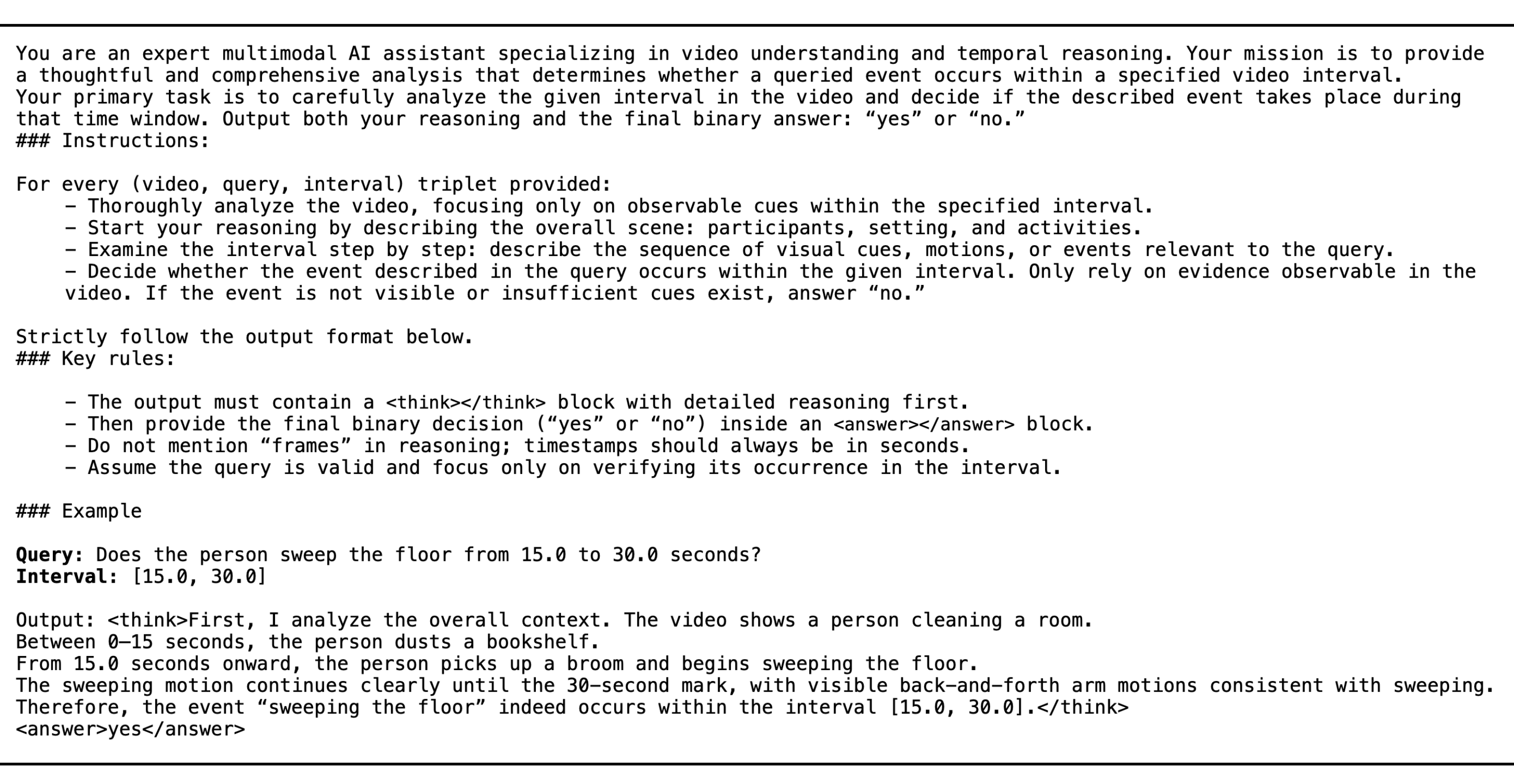}
    \caption{
    \textbf{Prompt for generating temporal reasoning for temporal verification.}
    }
    \label{fig: prompt_for_temporal_verification}
    \vspace{-3mm}
\end{figure}

\begin{figure}[t]
        \centering
        \includegraphics[width=1\linewidth]{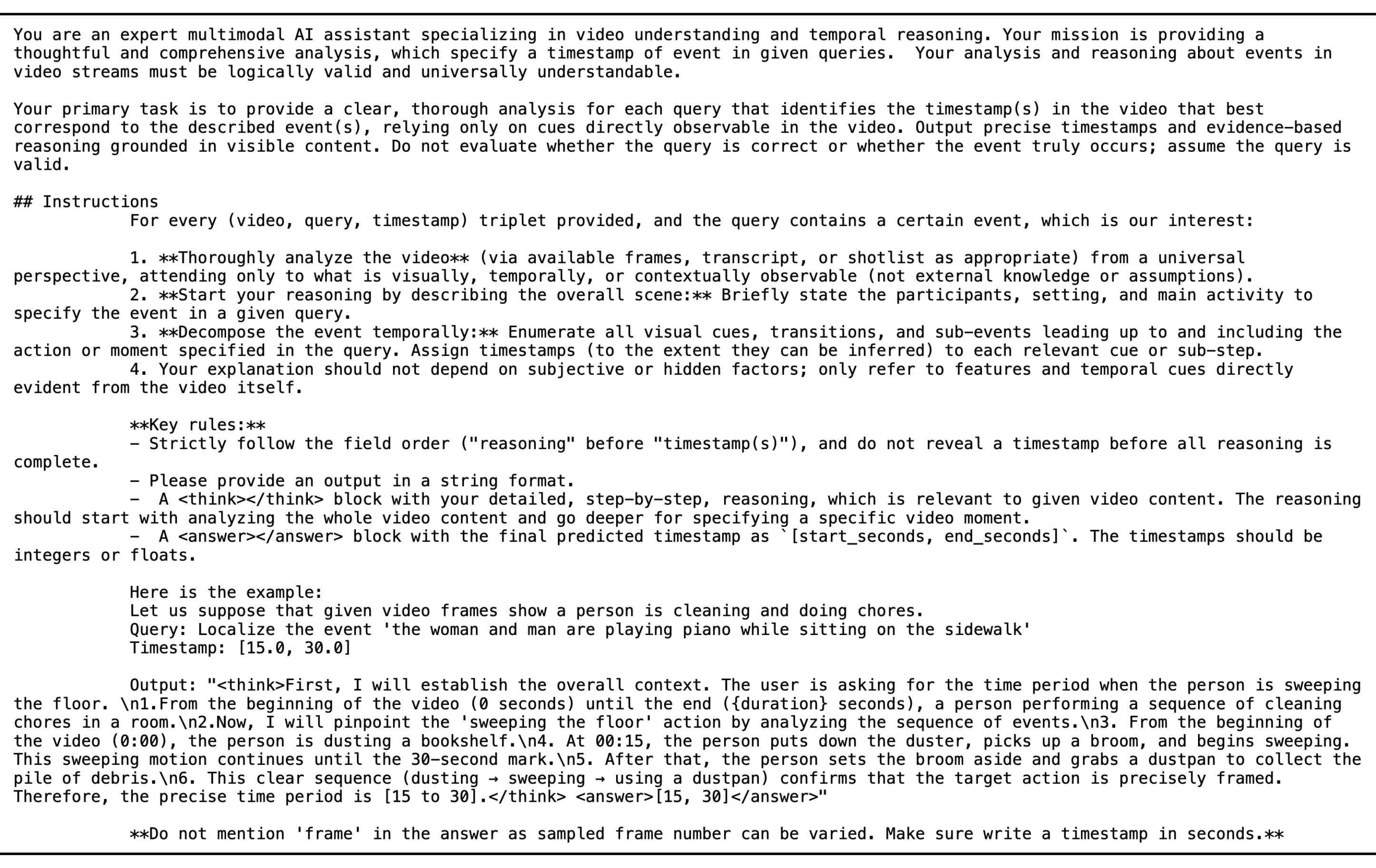}
    \caption{
    \textbf{Prompt for generating temporal reasoning for temporal grounding.}
    }
    \label{fig: prompt_for_video_reasoning_generation}
    \vspace{-3mm}
\end{figure}

To generate View30K, we design prompts in Fig.~\ref{fig: prompt_for_temporal_verification} and~\ref{fig: prompt_for_video_reasoning_generation} for each task. We sample frames every 1 second (\ie, 1 FPS) from videos and give them to GPT-5. We perform batch processing using GPT-API, and it involves less than 1 day for 3.6k videos. Additionally, the model fails to generate reasoning data for 0.3k videos, and a total of 3.3k with 61k reasoning for each task remains. 

For training, we set the max pixels for video processing to 2.8 M. We use the AdamW optimizer with a learning rate of 1e-6 and set the batch size to 8, 1 batch for each GPU. We set $\lambda$ as 0.7 in the reasoning reward.

\begin{table*}[t]
    \centering
        \caption{
            \textbf{Performance comparison between different training settings for a visual encoder.} In Vis. column, \snow~represents when we freeze the visual encoder, while \fire~denotes the visual encoder is trainable. All models are trained on both viewpoints. If there is an improvement, we use a blue color; otherwise, we use a red color.
        } 
        \vspace{-2mm}
        \resizebox{\linewidth}{!}{
        \begin{tabular}{lc cccccc cccccc}
        \toprule
        \bf \multirow{2}{*}{Methods} & \multirow{2}{*}{\bf Vis.} & \multicolumn{6}{c}{\bf CharadesEgo} & \multicolumn{6}{c}{\bf EgoExo-4D} \\ 
        \cmidrule(lr){3-8} \cmidrule(lr){9-14}
        & & V-Exo & V-Ego & V-ExoEgo & G-Exo & G-Ego & G-ExoEgo & V-Exo & V-Ego & V-ExoEgo & G-Exo & G-Ego & G-ExoEgo \\ \midrule
        \multirow{2}{*}{VideoChat2} 
        & \snow & 56.35 & 57.07 & 34.69 & 28.77 & 27.61 & 22.74 & 59.06 & 60.38 & 37.25 & 4.52 & 4.97 & 2.33 \\
        & \fire & \textcolor{red}{55.78} & \textcolor{red}{57.01} & \textcolor{red}{32.78} & \textcolor{blue}{29.76} & \textcolor{blue}{30.13} & \textcolor{blue}{23.86} & \textcolor{red}{58.32} & \textcolor{red}{54.55} & \textcolor{red}{31.25} & \textcolor{red}{2.13} & \textcolor{red}{2.63} & \textcolor{red}{1.13} \\
        \midrule
        \multirow{2}{*}{TimeChat} 
        & \snow & 52.49 & 48.59 & 27.91 & 60.30 & 61.79 & 46.51 & 51.64 & 50.31 & 28.33 & 10.57 & 11.48 & 4.33 \\
        & \fire & \textcolor{red}{50.76} & \textcolor{red}{45.11} & \textcolor{red}{21.64} & \textcolor{red}{55.87} & \textcolor{red}{59.76} & \textcolor{red}{42.57} & \textcolor{blue}{51.97} & \textcolor{blue}{51.34} & \textcolor{blue}{29.09} & \textcolor{red}{10.32} & \textcolor{blue}{11.76} & \textcolor{red}{4.23} \\
        \bottomrule
        \end{tabular}
        }
        \label{tbl: unfreeze}
\end{table*}

\begin{table*}[t]
    \caption{
    \textbf{Performance on EgoExo-Con across different subsets.} While performance for temporal verification remains steady across subsets, there is a noticeable performance gap between subsets in temporal grounding, likely due to the different difficulty of tasks. 
    }
    \vspace{-2mm}
    \resizebox{\linewidth}{!}{
      \begin{tabular}{lccc cccccc cccccc ccc}
      \toprule
       \multirow{2}{*}{\bf Methods} & \multicolumn{6}{c}{\bf CharadesEgo } 
       & \multicolumn{6}{c}{\bf EgoExo-4D} 
       & \multicolumn{6}{c}{\bf LEMMA} \\
    \cmidrule(lr){2-7} \cmidrule(lr){8-13} \cmidrule(lr){14-19} 
      & V-Exo & V-Ego & V-ExoEgo & G-Exo & G-Ego & G-ExoEgo & V-Exo & V-Ego & V-ExoEgo & G-Exo & G-Ego & G-ExoEgo & V-Exo & V-Ego & V-ExoEgo & G-Exo & G-Ego & G-ExoEgo \\ \midrule
      \textcolor{gray}{\textit{General-purpose}} \\ 
        VideoChat2 & 48.6 & 53.6 & 27.2 & 13.6 & 12.5 & 10.6 & 41.4 & 41.5 & 19.8 & 0.9 & 1.0 & 0.4 & 40.5 & 42.5 & 21.2 & 0.6 & 1.1 & 0.6 \\
        Qwen2.5-VL & 59.1 & 58.4 & \underline{42.2} & 31.4 & 23.1 & 16.1 & \underline{58.4} & \underline{59.3} & \textbf{49.5} & 5.3 & 5.3 & 2.7 & 55.6 & 59.2 & \underline{41.9} & 0.6 & 1.7 & 0.0 \\
        Video-LLaMA2 & 54.0 & 52.4 & 28.2 & 27.8 & 27.8 & 17.6 & 51.5 & 52.8 & 28.1 & 2.5 & 1.6 & 1.5 & 50.6 & 50.8 & 25.4 & 2.2 & 2.8 & 1.7 \\ 
        Video-LLaMA3 & \underline{61.8} & \underline{57.9} & 40.3 & \underline{57.1} & \underline{51.1} & \underline{37.3} & 52.6 & 51.5 & 33.6 & \underline{8.7} & \textbf{12.0} & 3.1 & 53.9 & \underline{59.5} & 36.3 & \textbf{11.7} & \textbf{22.9} & {5.6} \\ \midrule
      \textcolor{gray}{\textit{Time-aware}} \\
        VTimeLLM & 49.5 & 47.0 & 23.8 & 19.5 & 12.5 & 7.5 & 49.0 & 49.0 & 23.6 & 8.4 & \underline{10.1} & \textbf{6.3} & 46.4 & 51.7 & 23.5 & \underline{10.1} & 11.7 & \underline{6.1} \\
        TimeChat & 48.0 & 48.1 & 25.2 & 44.9 & 46.1 & 30.1 & 49.5 & 49.3 & 25.4 & 4.9 & 6.3 & \underline{3.3} & 49.2 & 46.4 & 23.5 & 5.0 & 5.0 & 3.4 \\
        TimeSuite & 47.7 & 47.6 & 24.7 & \textbf{63.4} & \textbf{56.2} & \textbf{44.8} & 45.8 & 47.3 & 25.2 & 5.8 & 8.3 & 2.3 & {54.0} & \underline{56.3} & 30.6 & 10.6 & \underline{16.2} & \textbf{6.7} \\
        TimeChat-VT & \textbf{64.6} & \textbf{63.7} & \textbf{44.8} & 56.6 & 54.2 & 35.8 & \textbf{60.4} & \textbf{59.4} & \underline{38.8} & \textbf{9.3} & 8.4 & \underline{3.3} & \textbf{60.9} & \textbf{65.1} & \textbf{45.5} & \underline{10.1} & {12.8} & 3.9 \\
  \bottomrule
  \end{tabular}
}
\label{tbl: performance across subsets}
\end{table*}

\section{Additional Experiments}

\subsection{The impact of Unfreezing Video Encoder}
In Table~\ref{tbl: unfreeze}, we conduct fine-tuning models: VideoChat2~\cite{li2024mvbench} and TimeChat~\cite{timechat} when fine-tuned with their video encoders unfrozen. Despite updating the visual encoders during training, we observe no further improvements; in fact, the models often underperform compared to the settings with frozen encoders. We conjecture that this may be due to overfitting or the limited scale of training data, which may not sufficiently support end-to-end tuning of large visual backbones. Furthermore, this supports the previous findings that naively mixing both viewpoints does not easily lead to improved view-invariant understanding.

\subsection{Performance on EgoExo-Con across subsets}
Table~\ref{tbl: performance across subsets} reports performance across models and subsets. The performance for temporal verification remains steady across subsets. In contrast, there is a noticeable performance gap between CharadesEgo and the others. Specifically, the models tend to struggle with accurate grounding for EgoExo-4D and LEMMA than CharadesEgo due to their lengthy videos and short moments, as shown in Fig.~\ref{fig: statistics}. Despite domain differences, we find consistent findings in Table~\ref{tbl: main}, significantly lagging in consistency compared to single-viewpoint performance.

\subsection{The impact of reasoning length}
\label{appendix: reasoning length}
\begin{wrapfigure}[15]{r}{0.6\textwidth}
    \centering
    \vspace{-0.5cm}
    \includegraphics[width=1.0\linewidth]{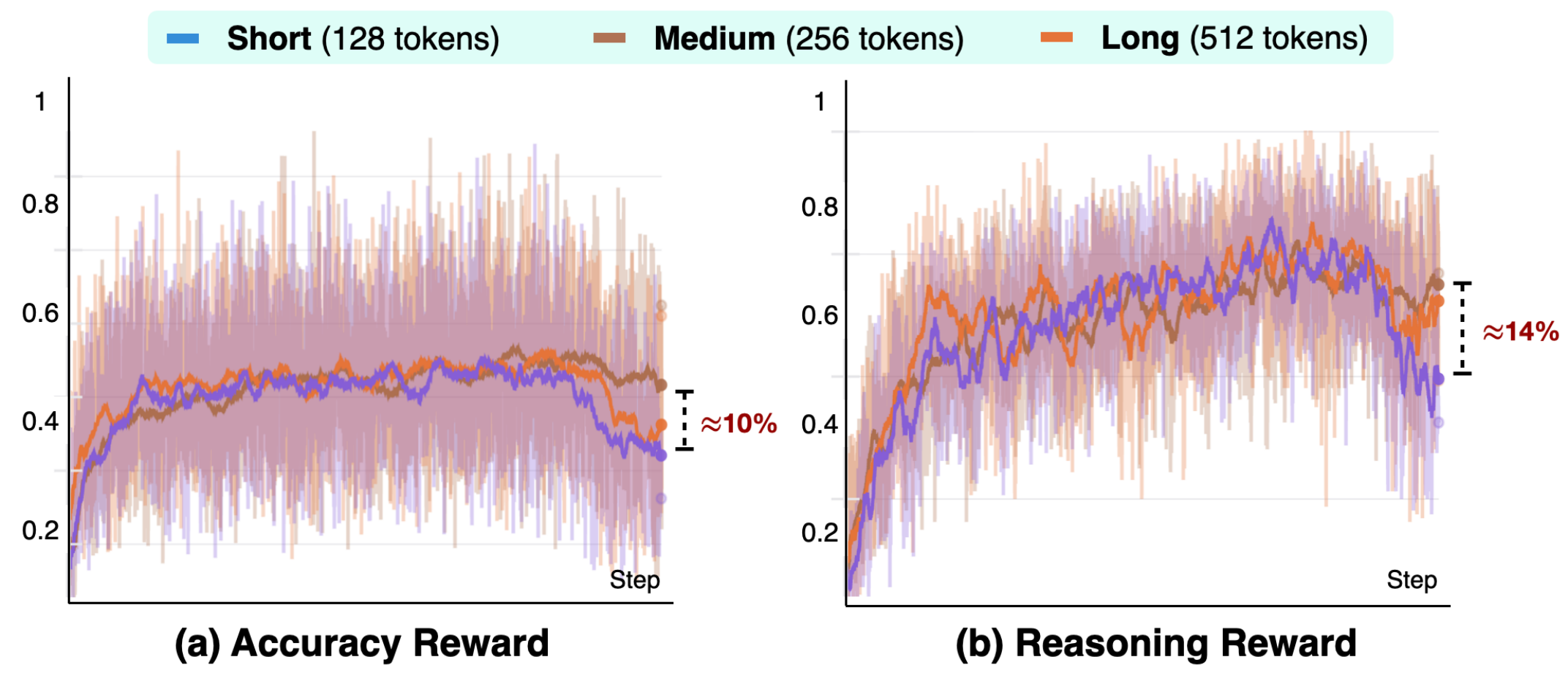}
    \caption{
   \textbf{The rewards across different reasoning lengths.} The medium-length reasoning shows the most stable rewards and balanced accuracy, suggesting that reasoning length critically affects optimization stability.
    }
    \label{fig: ablation2}
\end{wrapfigure}
We investigate how the length of the generated reasoning influences optimization stability and performance. Specifically, we categorize reasoning outputs into three groups: short (128 tokens), medium (256 tokens), and long (512 tokens), and train each configuration for 2k steps. Note that we chose the medium-length reasoning in View-GRPO. As shown in the Fig.~\ref{fig: ablation2}, the medium-length reasoning achieves the most stable rewards and balanced accuracy, while the short and long variants exhibit distinct issues: (1) Short reasoning tends to produce insufficient context, resulting in low accuracy and unstable optimization. (2) While long reasoning achieves high reasoning rewards, it often leads to lower accuracy, indicating over-exploration or hallucination. This also exposes a limitation of LLM-based judges, where higher rewards do not necessarily reflect factual correctness, as the judge may prioritize verbosity or perceived reasoning depth over actual accuracy. We also experimented with longer reasoning length (1024 tokens $<$), but did not observe further improvements; in fact, training became less stable and prone to over-exploration. These findings indicate that reasoning length critically affects optimization stability and that excessively short or long reasoning can be detrimental.

\section{Discussion}
\label{appendix: discussion}
For future directions, we outline three perspectives: data, architecture, and learning. From a data perspective, while large-scale ego–exo pre-training could eventually yield more view-invariant representations, collecting comprehensively synchronized multi-view data at scale is costly and often impractical.~\cite{dou2024unlocking} shows that egocentric-style data can be synthesized from exocentric videos by cropping or reprojecting hand–object regions. Such augmentations could help approximate cross-view signals without requiring paired capture. From an architecture perspective, current Video-LLM training paradigms process videos independently, making it difficult for models to perceive and compare viewpoints jointly. Architectures that enable simultaneous multi-video processing or explicit cross-view alignment signals could substantially improve consistency. From a learning perspective, prediction inconsistencies highlight shortcuts in the learning of view-specific biases. We believe that explicitly reinforcing rationale and temporal reasoning could benefit consistency, as shown via View-GRPO. 


\section{Additional Visualization}
\begin{wrapfigure}[10]{r}{0.5\textwidth}
    \centering
    \vspace{-1cm}
    \includegraphics[width=1.0\linewidth]{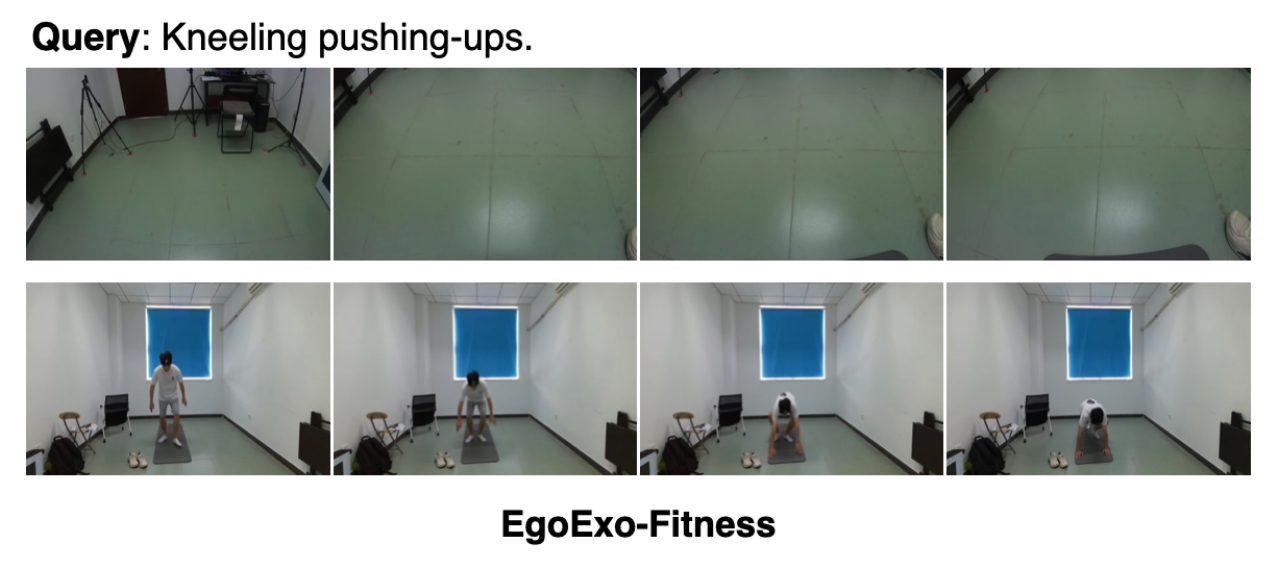}
    \caption{
   \textbf{A video sample from EgoExo-Fitnesses.} The given query is hard to identify in the egocentric video (top).
    }
    \label{fig: egoexo-fitnesse}
\end{wrapfigure}
\subsection{Unsuitable Case}
\label{appendix: unsuitable cases}
As we discussed in the dataset section, some videos are naturally unsuitable to include in our benchmark. In Fig.~\ref{fig: egoexo-fitnesse}, both videos are annotated with the query ``Kneeling pushing-ups.'' However, the egocentric view is insufficient to identify the action due to a limited field of view that only shows the ground, while the exocentric view clearly shows the full body. Due to this kind of noise, we conducted a strict curation and verification during the construction of EgoExo-Con.

\subsection{Model Responses Examples}
\begin{figure}[t]
        \centering
        \includegraphics[width=1\linewidth]{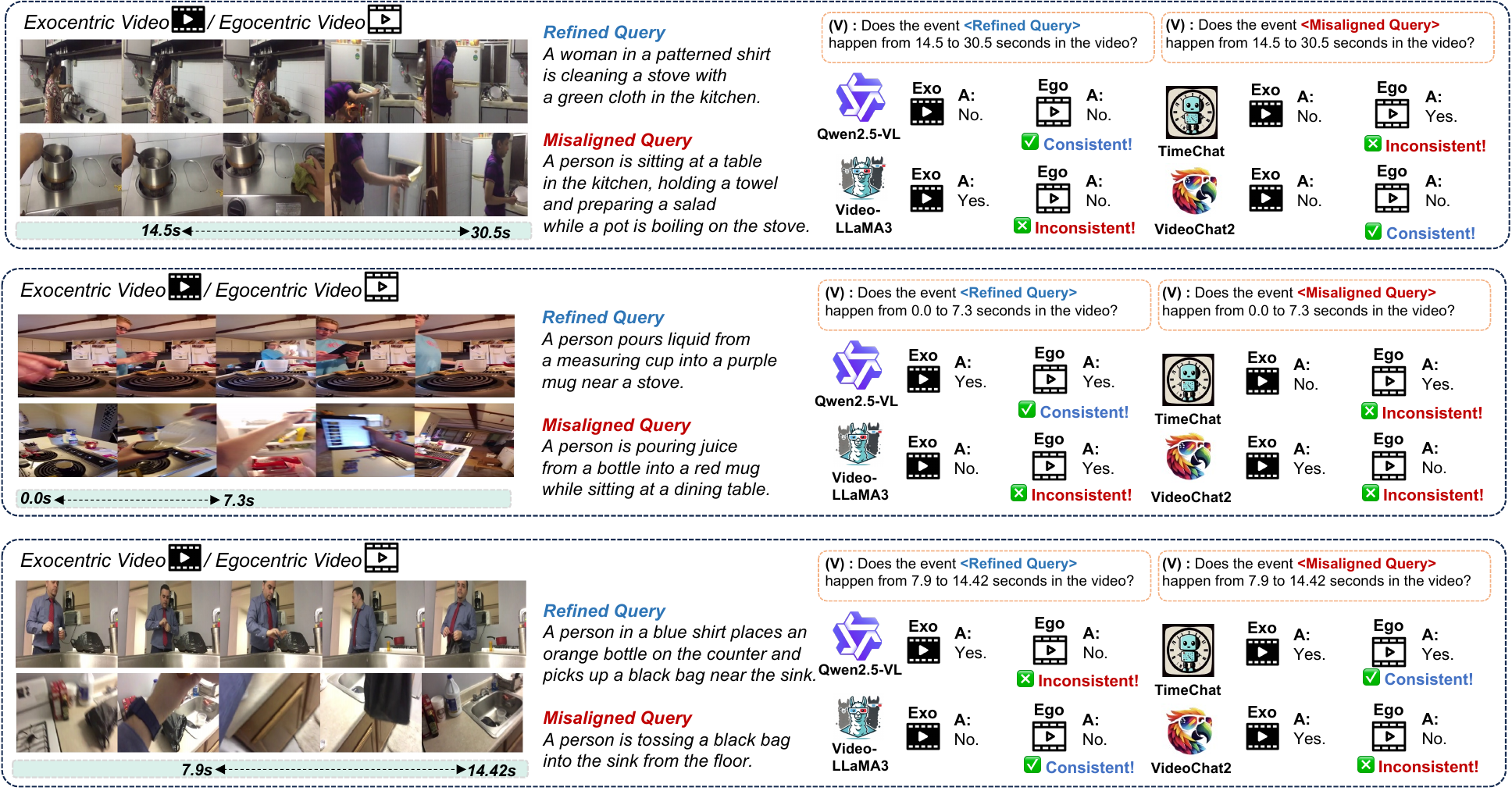}
    \caption{
    \textbf{Examples of test clips and model responses for temporal verification.}. Each row is from CharadesEgo and shows an exocentric--egocentric pair, a \textit{Refined Query} (positive) and a \textit{Misaligned Query} (negative), and per-view answers from each model. We mark consistency (\cmark) and inconsistency (\xmark) cases.
    }
    \label{fig: appdx_vis_verif}
\end{figure}

\begin{figure}[t]
        \centering
        \includegraphics[width=1\linewidth]{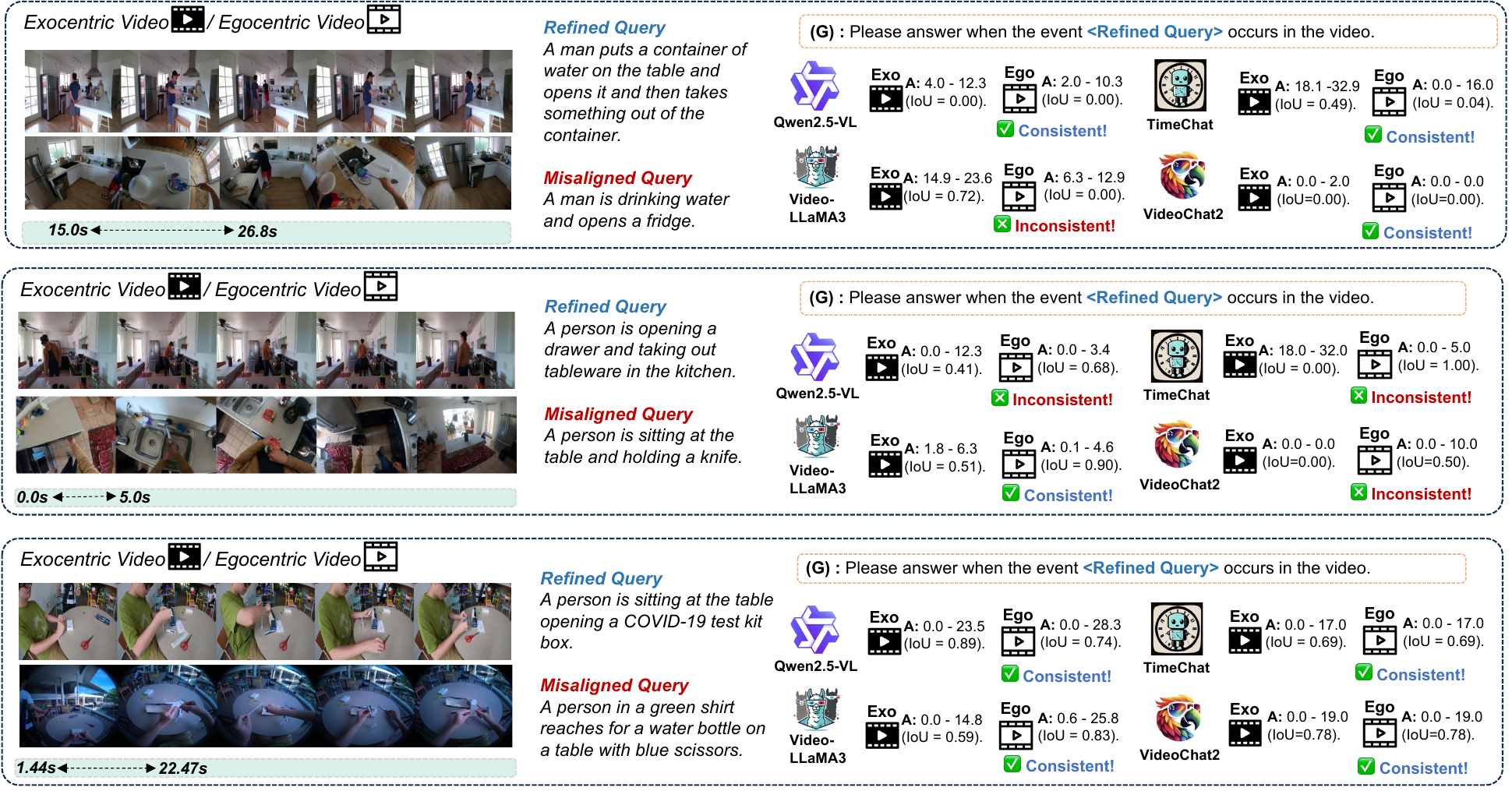}
    \caption{
    \textbf{Examples of test clips and model responses for temporal grounding}. The videos in the top two rows are from LEMMA, and the bottom row shows EgoExo4D videos. Each row presents an exocentric--egocentric pair with a \textit{Refined Query}; models output per-view time spans (with IoU to ground truth when available), and we indicate cross-view consistency (\cmark) vs.\ inconsistency (\xmark).
    }
    \label{fig: appdx_vis_grounding}
\end{figure}

We further provide model responses on EgoExo-Con across tasks. Fig.~\ref{fig: appdx_vis_verif} shows model responses for temporal verification from CharadesEgo, and Fig.~\ref{fig: appdx_vis_grounding} illustrates grounding predictions from LEMMA and EgoExo-4D. Note that we do not utilize misaligned queries for temporal grounding.

\begin{figure}[t]
        \centering
        \includegraphics[width=1\linewidth]{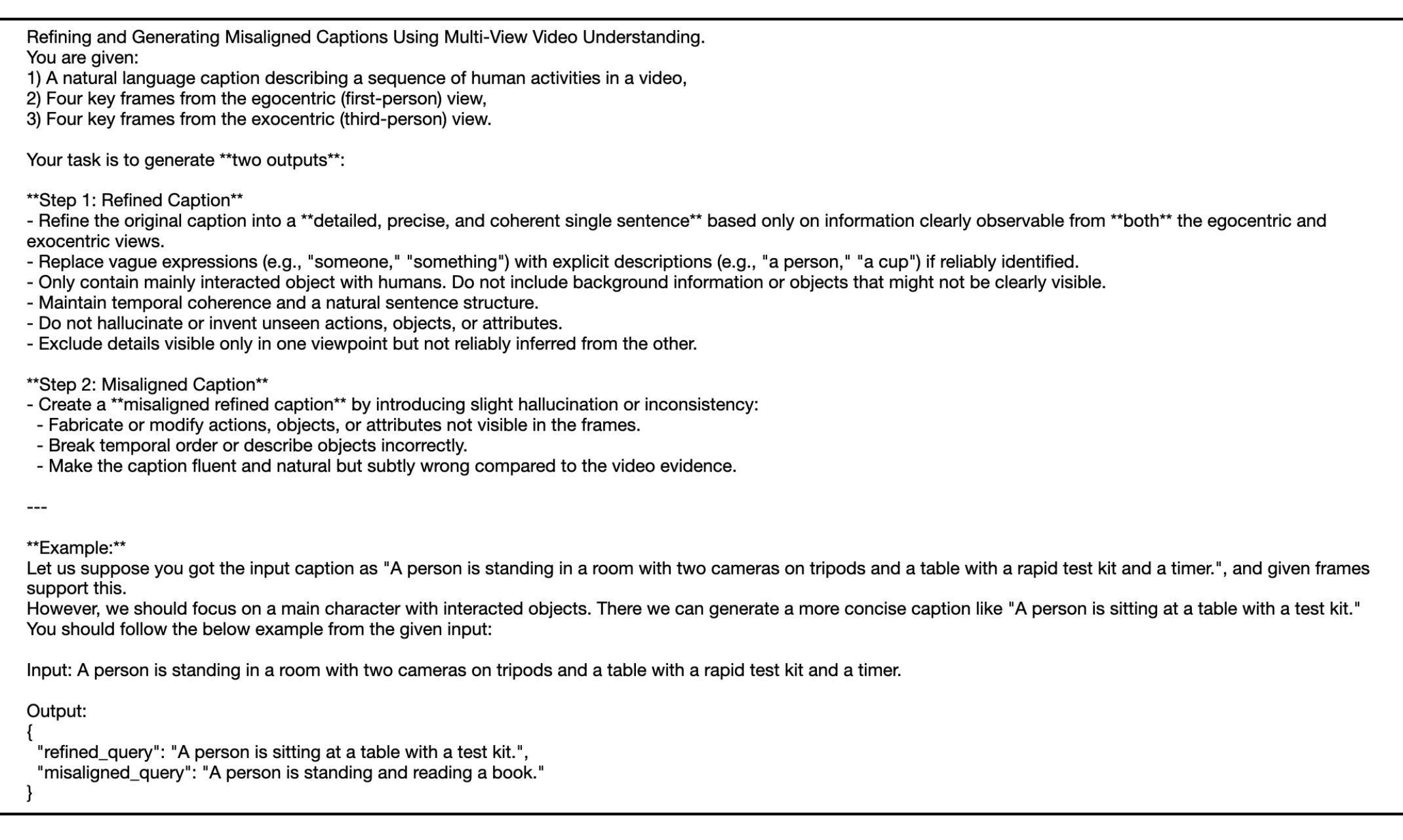}
    \caption{
    \textbf{Prompt for refined and misaligned queries.}
    }
    \label{fig: prompt_for_refinement}
    \vspace{-3mm}
\end{figure}

\begin{figure}[t]
        \centering
        \includegraphics[width=1\linewidth]{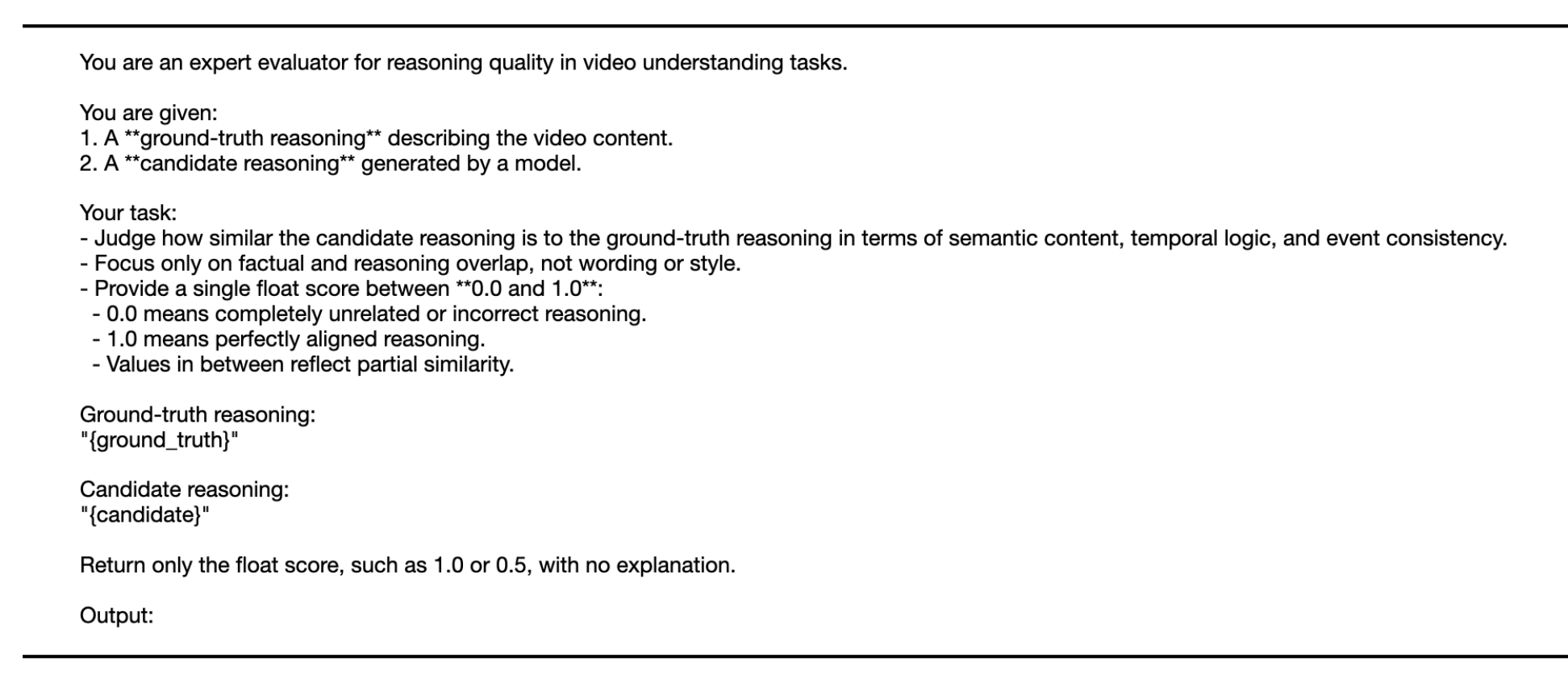}
    \caption{
    \textbf{Prompt for the reasoning reward function in View-GRPO.}
    }
    \label{fig: prompt_for_reward_reasoning}
    \vspace{-3mm}
\end{figure}

\section{Ethics review}
We confirm that our release will include only our annotations (\ie, refined query sentences and associated metadata) on top of the existing publicly available videos. Regarding annotator compensation, we clarify that all annotation and verification work was conducted by fairly compensated in-lab annotators.

\end{document}